\documentclass[twoside]{article}
\usepackage[accepted]{aistats2024}
\usepackage[utf8]{inputenc} 
\usepackage[T1]{fontenc}    
\usepackage{hyperref}       
\usepackage{url}            
\usepackage{multirow}       
\usepackage[ruled]{algorithm2e}    
\usepackage{booktabs}       
\usepackage{amsfonts}       
\usepackage{nicefrac}       
\usepackage{lipsum}
\usepackage{graphicx}       
\usepackage{amsmath}
\usepackage{amsthm}
\usepackage{bbm}
\usepackage[table,xcdraw]{xcolor}
\usepackage{amssymb}
\usepackage{pifont}
\usepackage[title]{appendix}
\usepackage{longtable}

\usepackage[round]{natbib}

\newcommand{\ba}{\mathbf{a}}
\newcommand{\x}{\mathbf{x}}
\newcommand{\X}{\mathbf{X}}
\newcommand{\V}{\mathbf{V}}
\newcommand{\bv}{\mathbf{v}}
\newcommand{\D}{\mathbf{D}}
\newcommand{\Z}{\mathbf{Z}}
\newcommand{\z}{\mathbf{z}}

\newcommand{\E}{\mathbb{E}}
\newcommand{\cmark}{\ding{51}}
\newcommand{\xmark}{\ding{55}}

\newtheorem{proposition}{Proposition}
\newtheorem*{theorem*}{Theorem}
\newtheorem{propositionrepeat}{Proposition}

\theoremstyle{remark}
\newtheorem{remark}{Remark}

\usepackage{enumitem}
\begin{document}

\runningtitle{Safe and Interpretable Estimation of Optimal Treatment Regimes}

\runningauthor{Parikh, Lanners, Akras, Zafar, Westover, Rudin, Volfovsky}
\twocolumn[

\aistatstitle{Safe and Interpretable Estimation of Optimal Treatment Regimes}

\aistatsauthor{Harsh Parikh* \And Quinn Lanners* \And Zade Akras \And Sahar F. Zafar}

\aistatsaddress{Johns Hopkins \\University \And  Duke University \And Harvard Medical School \And Massachusetts \\General Hospital} 

 \aistatsauthor{M Brandon Westover \And Cynthia Rudin \And Alexander Volfovsky}

 \aistatsaddress{ 
Beth Israel Deaconess \\Medical Center \And Duke University \And  Duke University} 

 ]
\begin{abstract}
Recent advancements in statistical and reinforcement learning methods have contributed to superior patient care strategies. However, these methods face substantial challenges in high-stakes contexts, including missing data, stochasticity, and the need for interpretability and patient safety. Our work operationalizes a safe and interpretable approach for optimizing treatment regimes by matching patients with similar medical and pharmacological profiles. This allows us to construct optimal policies via interpolation. Our comprehensive simulation study demonstrates our method's effectiveness in complex scenarios. We use this approach to study seizure treatment in critically ill patients, ultimately advocating for personalized strategies based on medical history and pharmacological features. Our findings recommend reducing medication doses for mild, brief seizure episodes and adopting aggressive treatment strategies for severe cases, leading to improved outcomes.
\end{abstract}
\section{INTRODUCTION}
Our study investigates optimal treatment strategies for critically ill patients suffering from seizures or epileptiform activity (EA). These conditions are associated with elevated in-hospital mortality rates and long-term disabilities \citep{parikh2023effects, ganesan2019electrographic, kim2018}. EA is commonly observed in patients with various medical conditions such as brain injuries \citep{lucke2015traumatic}, cancer \citep{lee2013risk}, organ failure \citep{boggs2002seizures}, and infections like COVID-19. Healthcare professionals in intensive care units (ICUs) frequently use anti-seizure medications (ASMs) to manage EA. However, there are concerns regarding the utilization of highly potent ASMs due to their potential adverse health effects \citep{farrokh2018antiepileptic, de2016effect}.  Additionally, the relative risks and benefits of ASMs vary among patients. This variation necessitates personalized treatment strategies to achieve optimal outcomes for each individual patient, as there is no one solution that fits all.\footnote{Strategies regarding when and how to treat patients based on their recent history are referred to as treatment regimes (denoted by $\pi_i$ for each patient $i$).}

We analyze data from a large hospital to identify optimal treatment regimes and generate clinically relevant hypotheses for future investigations in critical care. However, our data faces many challenges such as (i) a relatively small dataset of 995 patients, (ii) limited observation windows resulting in unobserved or missing ASM and EA data, and (iii) highly variable brain-drug interactions. No existing optimal treatment regime estimation methods are well-suited to handle these challenges (see Table~\ref{tab: lit-review}, Section \ref{sec: literature}, and Appendix \ref{sec: lit_survey_appendix}). While our study focuses on treating EA in critically ill patients, the underlying framework is applicable across various medical and healthcare contexts, such as addressing substance use disorder in intravenous drug users \citep{volkow2020personalizing}, managing coronary heart disease in ICU patients \citep{guo2022learning} or treating chronic psychiatric disorders \citep{murphy2007methodological}.

\textbf{Contributions.} We offer a general and flexible approach that allows for consistent estimation of optimal treatment regimes in the face of these challenges. Our approach is divided into three main steps:
\setlist{nolistsep}
\begin{enumerate}[noitemsep]
    \itemsep0em 
    \item \textbf{Pharmacological Feature Estimation:} We estimate patient-specific pharmacological features using a mechanistic model that captures EA-ASM interaction and is motivated by the underlying biochemistry. 
    \item \textbf{Distance Metric Learning:} We employ distance metric learning to identify clinical and pharmacological features affecting the outcome and use it to perform nearest-neighbors estimation to account for confounding factors. 
    \item \textbf{Optimal Regime Estimation:} We estimate the optimal treatment regime for each patient using their matched group. The matched group is comprised of nearby points according to the learned distance metric. The optimal regime is estimated using linear interpolation over the regimes of the nearby patients with favorable outcomes.
\end{enumerate}
Estimation via our approach results in personalized optimal treatment regimes that are:
\setlist{nolistsep}
\begin{itemize}[noitemsep]
    \itemsep0em 
    \item \textbf{\textit{Interpretable}}, allowing caregivers to understand, validate, and implement the regimes easily;
    \item \textbf{\textit{Safe}}, ensuring that patients are neither over-prescribed nor under-prescribed ASMs; and
    \item \textbf{\textit{Accurate}}, outperforming or performing on par with state-of-the-art black-box methods.
\end{itemize}
The simplicity and transparency of our approach coupled with its flexibility and interpretability makes it suited for high-stakes scenarios, such as the design of treatment strategies for patients experiencing epileptiform activity (EA) in the ICU. We discuss the identification of optimal treatment regimes in Section~\ref{sec: id} and delineate our methodology to estimate them in Section~\ref{sec: method}. We validate and compare our approach with existing methods via simulation studies in Section~\ref{sec: synth} and Appendix~\ref{sec: appendix_synth_exp}.

\textbf{Clinical Findings.} 
We show in Section~\ref{sec: seizure_study} that our estimated treatment regimes \textit{would have improved the outcomes for patients in the ICU}. The results indicate that a one-size-fits-all approach to escalating ASM usage in response to EA may not be universally beneficial. Instead, it is crucial to tailor treatment plans for each individual. For instance, patients exhibiting cognitive impairment or dementia are at a heightened risk of experiencing adverse effects from ASMs. A more cautious and lower-intensity approach to treatment may be warranted in such cases. This analysis not only characterizes beneficial approaches for treating EA in critically ill patients but also generates relevant hypotheses for future inquiry.

\section{RELATED LITERATURE}\label{sec: literature}
\begin{table}[]\label{tab: lit-review}
\resizebox{\linewidth}{!}
{\begin{tabular}{|
>{\columncolor[HTML]{D4D4D4}}l |
>{\columncolor[HTML]{FFCCC9}}c |
>{\columncolor[HTML]{9AFF99}}c |c|c|
>{\columncolor[HTML]{9AFF99}}c |c|}
\hline
\cellcolor[HTML]{B0B3B2}\textbf{Methods}   & \multicolumn{1}{l|}{\cellcolor[HTML]{B0B3B2}\textbf{CA}} & \multicolumn{1}{l|}{\cellcolor[HTML]{B0B3B2}\textbf{VT}} & \multicolumn{1}{l|}{\cellcolor[HTML]{B0B3B2}\textbf{MT}}  & \multicolumn{1}{l|}{\cellcolor[HTML]{B0B3B2}\textbf{LO}} & \multicolumn{1}{l|}{\cellcolor[HTML]{B0B3B2}\textbf{DE}} & \multicolumn{1}{l|}{\cellcolor[HTML]{B0B3B2}{\color[HTML]{333333} \textbf{IN}}} \\ \hline
\textbf{Our Method}                       & \cellcolor[HTML]{9AFF99}\cmark                           & \cmark                                                   & \cellcolor[HTML]{9AFF99}\cmark                            & \cellcolor[HTML]{9AFF99}\cmark                           & \cmark                                                   & \cellcolor[HTML]{9AFF99}\cmark                                                  \\ \hline
\textbf{Finite BI}                        & \xmark                                                  & \cellcolor[HTML]{FFCCC9}\xmark                          & \cellcolor[HTML]{FFCCC9}\xmark                           & \cellcolor[HTML]{9AFF99}\cmark                           & \cmark                                                   & \cellcolor[HTML]{FFCE93}$\Delta$                                              \\ \hline
\textbf{Infinite HZ}                      & {\color[HTML]{333333} \xmark}                                 & {\color[HTML]{333333} \cmark}                                 & \cellcolor[HTML]{FFCE93}{\color[HTML]{333333} $\Delta$} & \cellcolor[HTML]{FFCCC9}\xmark                          & \cmark                                                   & \cellcolor[HTML]{FFCE93}$\Delta$                                              \\ \hline
\cellcolor[HTML]{D4D4D4}\textbf{Censored DTR} & \xmark                                                  & \cmark                                                   & \cellcolor[HTML]{9AFF99}\cmark                            & \cellcolor[HTML]{FFCE93}$\Delta$                       & \cmark                                                   & \cellcolor[HTML]{FFCE93}$\Delta$                                              \\ \hline
\textbf{Deep RL}                          & \cellcolor[HTML]{9AFF99}\cmark                           & \cmark                                                   & \cellcolor[HTML]{FFCE93}$\Delta$                        & \cellcolor[HTML]{FFCE93}$\Delta$                          & \cellcolor[HTML]{FFCE93}$\Delta$                           & \cellcolor[HTML]{FFCCC9}\xmark                                                 \\ \hline
\textbf{Causal NN}                        & \xmark                                                  & \cellcolor[HTML]{FFCCC9}\xmark                          & \cellcolor[HTML]{FFCCC9}\xmark                           & \cellcolor[HTML]{9AFF99}\cmark                           & \cmark                                                   & \cellcolor[HTML]{9AFF99}\cmark                                                  \\ \hline
\end{tabular}}
\caption{Characteristics of optimal regime estimation approaches. \textit{Finite BI}: finite timestep backward induction methods, \textit{Infinite HZ}: infinite horizon methods, \textit{Censored}: censored data methods, \textit{Deep RL}: deep reinforcement learning methods, \textit{Causal NN}: causal nearest neighbors. Columns represent \textit{CA}: continuous action space, \textit{VT}: variable timesteps, \textit{MT}: missing timesteps, \textit{LO}: targets long-term outcomes without requiring a designed reward function, \textit{DE}: data efficiency, and \textit{IN}: interpretability. Green cells denote desired properties and red cells indicate undesired properties in the context of our problem. $\Delta$ indicates the attribute depends on underlying modeling choices.}
\end{table}

Our literature survey encompasses various techniques for estimating optimal treatment regimes. We classify these techniques into five categories: (i) Finite Timestep Backward Induction \citep{murphy2003optimal, robins2004optimal, murphy2005generalization, moodie2010estimating, chakraborty2010inference, zhang2012estimating, zhao2015new, murray2018bayesian, blumlein2022learning, qian2011performance, moodie2014q, zhang2018interpretable}, (ii) Infinite Time Horizon \citep{ernst2005tree, ertefaie2018constructing, clifton2020q}, (iii) Censored Data \citep{goldberg2012q, lyu2023imputation, zhao2020constructing}, (iv) Deep Reinforcement Learning \citep{mnih2013playing, lillicrap2015continuous, haarnoja2018soft, fujimoto2018addressing, kumar2020conservative, fujimoto2018off, wang2020critic}, and (v) Causal Nearest Neighbors \citep{zhou2017causal}. 

Each category of techniques has its strengths and limitations. Finite timestep backward induction methods offer interpretability and ease of implementation. However, they struggle with missing states, samples with variable timesteps, and large action spaces. Infinite time horizon and censored data methods can handle more nuanced temporal data but require a predefined reward function. Deep reinforcement learning (RL) can handle more complex regimes but lacks interpretability and requires a large sample size. There is a need for a method that can handle continuous state and action spaces, variable and missing timesteps, does not require the specification of an arbitrary reward function, and can work with a small sample size while maintaining accuracy and interpretability. We provide a summary of each category of techniques in regard to these attributes in Table~\ref{tab: lit-review} and include an in-depth literature survey in Appendix~\ref{sec: lit_survey_appendix}.

\section{PRELIMINARIES}\label{sec: prelim}
We now introduce our setup and notation. While our study focuses on treating EA in critically ill patients, the underlying framework is applicable across various medical and healthcare contexts, as discussed earlier.

For each patient $i$ in a cohort of $n$ patients, we observe (i) pre-treatment covariates $\X_i$, (ii)
time-series of states $\{E_{i,t}\}_{t=1}^{T_i}$ (in this case the EA burden), where $T_i$ is the duration for which the patient is under observation, (iii) sequence of actions $\{\Z_{i,t}\}_{t=1}^{T_i}$ (a vector of ASM drug doses given to the patient), and (iv) discharge outcome $Y_i$. Here, $Y_i$ is a binary indicator for patient well-being with $1$ indicating an adverse outcome based on the modified Rankin Score (mRS). The mRS was retrospectively abstracted from hospital records, specifically physician and physical therapy notes, at the time of patient discharge. 
The mRS assessments underwent rigorous independent review by evaluators, who were intentionally blinded to the patients' EEG measurements and antiseizure medication status to avoid bias.  

The sequence of actions, $\{\Z_{i,t}\}$, are determined based on the administered policy $\pi_i$ such that $\Z_{i,t} = \pi_i(\{E_{i,t'}\}_{t'=1}^{t},\{\Z_{i,t'}\}_{t'=1}^{t-1}) + \mathcal{E}_{i,t}$ where $\mathcal{E}_{i,t}$ is the unobserved time-and-patient specific factor affecting the action at time $t$. $Y_i(\{\z_{i,t}\})$ denotes the potential outcome, under the action sequence $\{\z_{i,t}\}$. However, since $\z_{i,t}$'s are determined by the policy $\pi_a$, we redefine the potential outcomes as a function of the policy itself, denoted as $Y_i(\pi_a)$. We assume that the observed outcome $Y_i$ is equal to the potential outcome under the administered treatment regime, denoted $Y_i(\pi_i)$. Note that while we observe $\Z_{i,t}$'s, we do not observe the underlying treatment regime $\pi_i$.

Our goal is to identify an optimal regime $\pi^*$ for each patient $i$ that minimizes their potential outcome: 
$$
\pi_i^* \in \arg \min_{\pi_a} \mathbb{E}[ Y_i(\pi_a) | \X_i].
$$ 
What makes this challenging is that we only observe the potential outcome corresponding to the treatment regime administered by the doctors. Thus, $Y_i =  Y_i(\pi_i)$ for each patient while all the other potential outcomes are missing (or unknown). Importantly, the outcome is observed at a timepoint $\tau_i$ which may be substantially further down the road than the length of observation for each patient, denoted by $T_i$. 

To address this missingness, we note that the state-action interaction and state transition are determined by underlying pharmacology that can be decoupled into two parts: (i) pharmacokinetics and (ii) pharmacodynamics. Pharmacokinetics describes the changes in drug concentration at time $t$ as a function of the drug concentration at the previous time points along with the current drug dose at time $t$. 
Pharmacodynamics describes the changes to the EA burden at time $t$ as a function of the current drug concentration and the past EA burden. 
The pharmacokinetic-pharmacodynamic (PK/PD) system is formalized as a pair of partial differential equations (described in detail in Appendix~\ref{sec: appendix_pkpd}). Since this structural system fully governs the drug-EA interaction, conditioning on it allows us to avoid complex outcome simulators while also providing context for the observed heterogeneity in outcomes.
\section{IDENTIFICATION}\label{sec: id}
We now discuss the underlying assumptions that allow identification of $\pi^*_i \in \arg \min_{\pi_a} \mathbb{E}[ Y_i(\pi_a) \mid \X_i]$ for each patient $i$. We start by assuming conditional ignorability \citep{rubin1974estimating, robins2000robust}, $Y_i(\pi_a) \perp \pi_i \mid \X_i$, an assumption standard in observational causal studies. This assumption is reasonable in our setting as we know that caregivers decide the drug regimes primarily based on the pre-treatment features $\X$. By the law of iterated expectations, we know that

\begin{equation*}
    \mathbb{E}[ Y_i(\pi_a) \mid \X_i] = \sum_\pi \begin{pmatrix}
        \mathbb{E}[ Y_i(\pi_a) \mid \X_i, \pi_i = \pi] \\\times P(\pi_i = \pi \mid \X_i)
    \end{pmatrix}.
\end{equation*}
And by conditional ignorability, 
\begin{equation*}
    \mathbb{E}[ Y_i(\pi_a) \mid \X_i, \pi_i = \pi] = \mathbb{E}[ Y_i(\pi_a) \mid \X_i, \pi_i = \pi_a]
\end{equation*} 
for all $\pi$. Thus, if we have positivity, i.e. $P(\pi_i = \pi_a \mid \X_i) > 0$, then $\mathbb{E}[ Y_i(\pi_a) \mid \X_i]$ is identifiable as $\mathbb{E}[ Y_i \mid \X_i, \pi_i = \pi_a]$.

There are many scenarios similar to our setting where the dimensionality of $\pi$ is high and experts' treatment choices are based on patients' characteristics. In these scenarios, it is \textit{highly unlikely} that $P( \pi_i = \pi \mid \X_i) > 0$ for all $\pi$ and $\X_i$. However, recall that we are particularly interested in identifying the optimal treatment regime $\pi^*_i$ for each patient $i$ and not identifying $\mathbb{E}[ Y_i(\pi_a) \mid \X_i]$ for any arbitrary policy $\pi_a$. Thus, for our context, it is reasonable to assume that even if the clinicians' policies are suboptimal, they are sampled from the neighborhood of the optimal policy such that $P(\pi_i = \pi^*_i \mid \X_i) = \mathbb{E}_{\pi \mid \X_i}[P(\pi_i = \pi \mid \X_i)]$. We refer to this assumption as \textit{local} positivity. This assumption is weaker than the standard positivity assumption in causal inference. 
The major implication of this assumption is that $P(\pi_i = \pi^*_i \mid \X_i) > 0$, allowing us to identify $\mathbb{E}[ Y_i(\pi^*_i) \mid \X_i]$ and subsequently $\pi^*_i = {\arg \min}_{\pi \text{ s.t. } P(\pi \mid \X_i)>0} \mathbb{E}[ Y_i \mid \X_i, \pi_i = \pi]$. 

Under the assumption of local positivity, if $\pi_i$ were observed for each patient $i$, $\pi^*_i$ is always identifiable. However, as noted in Section~\ref{sec: prelim}, we only observe $\{E_{i,t}\}_{t=1}^{T_i}$ and $\{\Z_{i,t}\}_{t=1}^{T_i}$ while the underlying $\pi_i$ is unobserved. Recall that $\Z_{i,t} = \pi_i(\{E_{i,t'}\}_{t'=1}^{t},\{\Z_{i,t'}\}_{t'=1}^{t-1}) +  \mathcal{E}_{i,t}$, where $\mathcal{E}_{i,t}$ is an unobserved patient-and-time specific factor. We make a Markovian assumption, $\pi_i(\{E_{i,t'}\}_{t'=1}^{t},\{\Z_{i,t'}\}_{t'=1}^{t-1}) = \pi_i(\{E_{i,t'}\}_{t'=t-12h}^{t},\{\Z_{i,t'}\}_{t'=t-12h}^{t-1})$
and a sequential ignorability assumption such that
$\mathcal{E}_{i,t}
    \perp 
    (\{E_{i,t'}\}_{t'=1}^{t},
    \{\mathcal{E}_{i,t'}\}_{t'=1}^{t-1})
    \mid \X_i.$
Under these assumptions, $\pi_i$ is non-parametrically identifiable for each patient $i$ \citep{matzkin2007nonparametric}. This, in turn, implies that the optimal treatment regime $\pi^*_i$ is identifiable.

\begin{remark}
Recall that the outcome $Y_i$ is a function of a high-dimensional vector of EA burdens $\{E_{i,t}\}_{t=1}^{\tau_i}$ and drug doses $\{ \Z_{i,t}\}_{t=1}^{\tau_i}$, some of which are unobserved. Defining the treatment as a regime $\pi_i$ is akin to exposure mapping such that even though $(\{E_{i,t}\}_{t=1}^{\tau_i},\{ \Z_{i,t}\}_{t=1}^{\tau_i}) \neq (\{E_{j,t}\}_{t=1}^{\tau_j},\{ \Z_{j,t}\}_{t=1}^{\tau_j})$ we have $\E[Y_i(\pi_i) | \X_i = \x] = \E[Y_j(\pi_j) | \X_j = \x]$ if $\pi_i = \pi_j$. This helps us address the problem with missing $E_{i,t}$'s and $\Z_{i,t}$'s and ensures that the local positivity assumption is more reasonable.
\end{remark}

\section{METHODOLOGY}\label{sec: method}
We now outline our three-stage methodology for estimating the optimal treatment regime. The first stage involves estimating an individualized mechanistic model from observed state-action data to approximate state transition dynamics. Mechanistic modeling offers interpretability and needs much less data for fine-tuning. 
We also estimate the administered regimes ($\pi_i$'s) if they are unobserved (as in our setup). In the second stage, we create a distance metric to match patients based on pre-treatment covariates and estimated mechanistic model parameters.
Subsequently, we use the estimated distance metric to tightly match patients 
. Finally, in the third stage, we leverage these matched groups to estimate the optimal treatment regimes. 

Our interpretable matching approach allows validation through case-based reasoning, which enhances confidence in the estimation procedure and underlying assumptions. We provide details for each stage in the following subsections, with a focus on our real-world application. However, the framework is adaptable to other applications with similar data structures.

\paragraph{Mechanistic State Transition Modeling.} We approximate PK using a one-compartment model \citep{shargel1999applied}, with half-life as the parameter, and Hill's PD model \citep{hill1910possible, weiss1997hill, nelson2008lehninger}, with receptor-ligand affinity and drug dose for 50\% efficacy as parameters, to model the short-term effectiveness of the ASMs in reducing EA burden. We delineate the models formally in Appendix~\ref{sec: appendix_pkpd}. 
For each patient $i$ in the cohort, we estimate these individualized PK/PD parameters by minimizing the mean squared error between the predicted EA time series under the observed ASM regime using the mechanistic model and the actual observed EA time series.
This step is akin to estimating a multi-dimensional propensity score.
\begin{remark}
We approximate state-transition dynamics via deterministic mechanistic models, but we do not use them for counterfactual simulations. Mechanistic modeling isolates clinically relevant pharmacological features from stochastic dynamics. While state-transition dynamics adjustment is not necessary for consistent estimation, accounting for PK/PD parameters aids in estimating heterogeneous effects, akin to using propensity scores with Bayesian regression trees  \citep{hahn2020bayesian}.
\end{remark}

\paragraph{Characterizing Administered Policies.} 
In our study, we focus on treatment regimes for two commonly used anti-seizure medications (ASMs): propofol and levetiracetam. For our application, we employ the policy template that is defined by the drug administration protocols used in hospitals, to ensure interpretability, although our framework can accommodate non-parametric policy functions such as trees or forests. Propofol, a sedating ASM, is administered as a continuous infusion based on the past 1hr, 6hrs, and 12hrs of seizure levels using policy $\pi^{prop}$. In contrast, non-sedative ASM levetiracetam is given as a bolus every 12 hours, with dosages varying according to recent EA burden and drug history through policy $\pi^{lev}$. The regime for patient $i$ is denoted by $$\pi_i = \begin{Bmatrix} 
\pi^{prop}_{i}\left(\{E_{i,t'}\}_{t'=1}^{t},\{\Z_{i,t'}\}_{t'=1}^{t-1} ; \ba^{p}_i \right) \\
\pi^{lev}_{i}\left(\{E_{i,t'}\}_{t'=1}^{t},\{\Z_{i,t'}\}_{t'=1}^{t-1} ; \ba^{l}_i \right)\}
\end{Bmatrix}.$$ We provide the functional forms of the policies in Appendix~\ref{sec: policy}. We use the observed EA burdens ($\{E_{i,t}\}$) and ASM doses ($\{\Z_{i,t}\}$) to deduce the administered policy $\pi_i$ for each patient $i$ by minimizing the mean squared error loss between the predicted and observed drug doses at each time $t$.  

\paragraph{Distance Metric Learning and Matching}
To adjust for confounding, we need to account for pre-treatment covariates and pharmacological features. We do this by grouping patients who are similar in these features but are treated differently. This procedure is called matching, a commonly used approach to nonparametrically estimate potential outcomes \citep{ho2007matching, stuart2010matching, parikh2022malts}. For the sake of simplicity, let $\V_i$ denote a vector of pre-treatment and pharmacological features for each patient $i$. Then, the estimate for $\mathbb{E}[ Y(\pi_a) | \V = \bv]$ is given by $\widehat{Y}_{\bv}(\pi_a) = m(MG_{d}(\mathcal{D}, r, \bv),\pi_a)$ where $MG_{d}(\mathcal{D}, r, \bv)$ is the matched group of units from dataset $\mathcal{D}$ that are $r$ distance away from $\bv$ under distance metric $d$, and $m$ is a regression on the units in the matched group evaluated at $\pi_a$. 

In high-dimensional scenarios with limited data, it is not possible to precisely match all covariates. Thus, we want to match tightly on important covariates that affect patients' prognoses. Recent matching approaches have explored distance metric learning before matching for more accurate and interpretable causal effect estimation \citep[][see Appendix~\ref{sec: appendix_matching} for further details]{parikh2022malts, diamond2013genetic, lanners2023variable}.
We extend the Variable Importance Matching (VIM) framework \citep{lanners2023variable} to our problem setting. Our distance metric $d$ is parameterized by a positive semi-definite matrix ${\mathcal{M}}$ such that $d_{\mathcal{M}}(\bv_i,\bv_k) = (\bv_i - \bv_k)^T \mathcal{M} (\bv_i - \bv_k)$. We constrain $\mathcal{M}$ to a diagonal matrix, enabling domain experts to interpret these entries as feature importance values. Consequently, we set $\mathcal{M}_{j,j}$ equal to the gini impurity importance of the $j$-th feature in the model for $\E[Y | \V ]$ (as defined in \citet{nembrini2018revival} and \citet{ishwaran2015effect}). To ensure the ``honesty'' of our approach, we split the dataset $\mathcal{D}$ into two parts: the training set $\mathcal{D}_{tr}$ and the estimation set $\mathcal{D}_{est}$ \citep{Ratkovic2019RehabilitatingTR}. We fit gradient-boosting trees with 100 estimators on $\mathcal{D}_{tr}$, each with a maximum depth of 2. Henceforth, we denote the learned distance metric as $d^\dagger$.

\paragraph{Estimating Optimal Regimes.} For each matched group centered around patient $i \in \mathcal{D}_{est}$, we consider the administered regimes $\pi_k$ and outcomes $Y_k$ for all $k \in MG_{d^\dagger}(\mathcal{D}_{est}, r, \V_i)$, where $d^\dagger$ is the learned distance metric. For the sake for simplicity, we will denote $MG_{d^\dagger}(\mathcal{D}_{est}, r, \V_i)$ as $MG_i$. We estimate the conditional expected outcome ${\nu}_{i}(\pi):=\E[Y_i \mid \pi, \V_i]$ using only the units in $MG_i$. The estimate is denoted as $\widehat{\nu}_{i}(\pi)$. Further, consider a \textit{new} operator $\bigoplus$ such that if $\pi_1 \in \mathrm{Dom}(\pi)$ (a function that maps states to a vector of ASM doses) and $\pi_2 \in \mathrm{Dom}(\pi)$ (another function that maps states to a vector of ASM doses) then $\pi_3 = \pi_1 \bigoplus \pi_2 \in \mathrm{Dom}(\pi)$. This operation is defined so that if $\pi_3 = \pi_1 \bigoplus \pi_2$ then $\pi_3(s) := \pi_1(s) + \pi_2(s)$ for all $s$ in the domain of states. Then, our estimate of the optimal treatment regime for unit $i$ is 
$\widehat{\pi}_i^* \in \arg \min_{\pi_{c,i}}\widehat{\nu}_i(\pi_{c,i}) \text{ where, } \pi_{c,i} = \underset{k \in MG_i}{\bigoplus} c_{k} \pi_k,$
$\sum_{k \in MG_i} c_k = 1$ and $0 \leq c_k \leq 1$.

\paragraph{Consistency.} We now discuss a smoothness of outcomes assumption under which our estimated optimal regime is consistent. Let's first define an $(\mathcal{S},p)$-norm on the space of policies such that $\|\pi_1 - \pi_2\|_{\mathcal{S},p} = \left(\int_{s\in\mathcal{S}} \left|\pi_1(s)-\pi_2(s)\right|^p\right)^{1/p} ds$ where $\mathcal{S}$ is the state space for the policies and $p$ is some positive integer. The smoothness of outcomes assumption is given as follows: given constants $\lambda_\pi\geq0$ and $\lambda_\V\geq0$ such that for any two units $1$ and $2$, if $\|\pi_1 - \pi_2\|_{\mathcal{S},\infty} \leq \lambda_\pi$ and $\|\V_1 - \V_2\|_2 \leq \lambda_\V$ then $\| \E[Y(\pi_1)\mid \V_1] - \E[Y(\pi_2)\mid \V_2] \| \leq \delta(\lambda_\pi,\lambda_\V)$ where $\delta$ is a monotonically decreasing function in both the arguments with $\delta(0,0) = 0$.
 
This assumption essentially implies that if $\V_1$ and $\V_2$ are close and if $\pi_1$ and $\pi_2$ are also close then the expected potential outcomes are also close.

\begin{proposition}\label{prop:consistent}
    Given the conditional ignorability, local positivity, and smoothness of outcomes assumptions, $\widehat{\pi}^*_i$ is a consistent estimate of $\pi^*_i$, such that
    \begin{equation*}
        \lim_{n\to\infty}\E[Y(\widehat{\pi}^*_i) \mid \V_i] \to \E[Y({\pi}^*_i) \mid \V_i].
    \end{equation*}
\end{proposition}
We provide the proof of this proposition in Appendix~\ref{sec: appendix_theorem}

\begin{remark}
As our regimes $\pi$ are linear score functions with parameter vector $\ba$ (see Appendix~\ref{sec: policy}), $\pi_{k_3} = \pi_{k_1} \bigoplus \pi_{k_2}$ corresponds to defining new policy $\pi_{k_3}$ with parameters $\ba_{k_3} = \ba_{k_1} + \ba_{k_2}$. This property comes in handy when comparing the administered policy's parameters with the estimated optimal policy.
\end{remark}
\section{SYNTHETIC EXPERIMENTS}\label{sec: synth}
\begin{figure*}[h]
    \centering
    \includegraphics[width=0.8\linewidth]{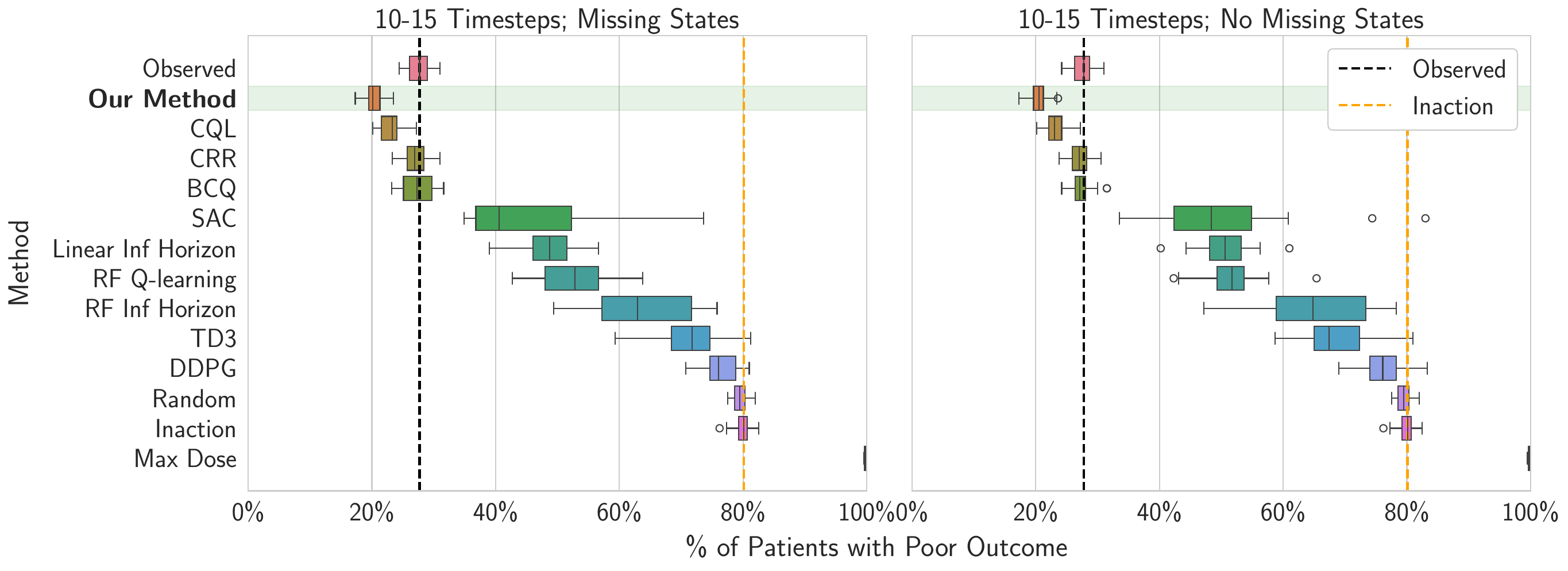}
    \caption{Percent of patients with poor outcomes under each method's proposed policy (\textit{lower is better}). Boxplots show the distribution of the average outcomes over 20 iterations. \textit{Observed} shows average observed outcomes. \textit{Inaction} and \textit{Max Dosing} administer no drugs and the max amount of drugs to each patient at each timestep, respectively. \textit{RF Q-learning} is a finite timestep backward induction method using random forests. \textit{Infinite (Inf) Horizon} methods use fitted Q-iteration \citep[see][]{clifton2020q} with either linear models or random forests. \textit{Q-learning} and \textit{Inf Horizon} discretize the treatment into five bins. \textit{BCQ, CQL, CRR, GGPQ, SAC}, and \textit{TD3} are Deep RL methods. \textit{Inf Horizon} and Deep RL methods use an insightful reward function, see Appendix~\ref{sec: appendix_comp_method}.}
    \label{fig: synthetic-sim1}
\end{figure*}

\begin{figure*}
    \centering
    \begin{tabular}{cc}
    \includegraphics[width=0.39\linewidth]{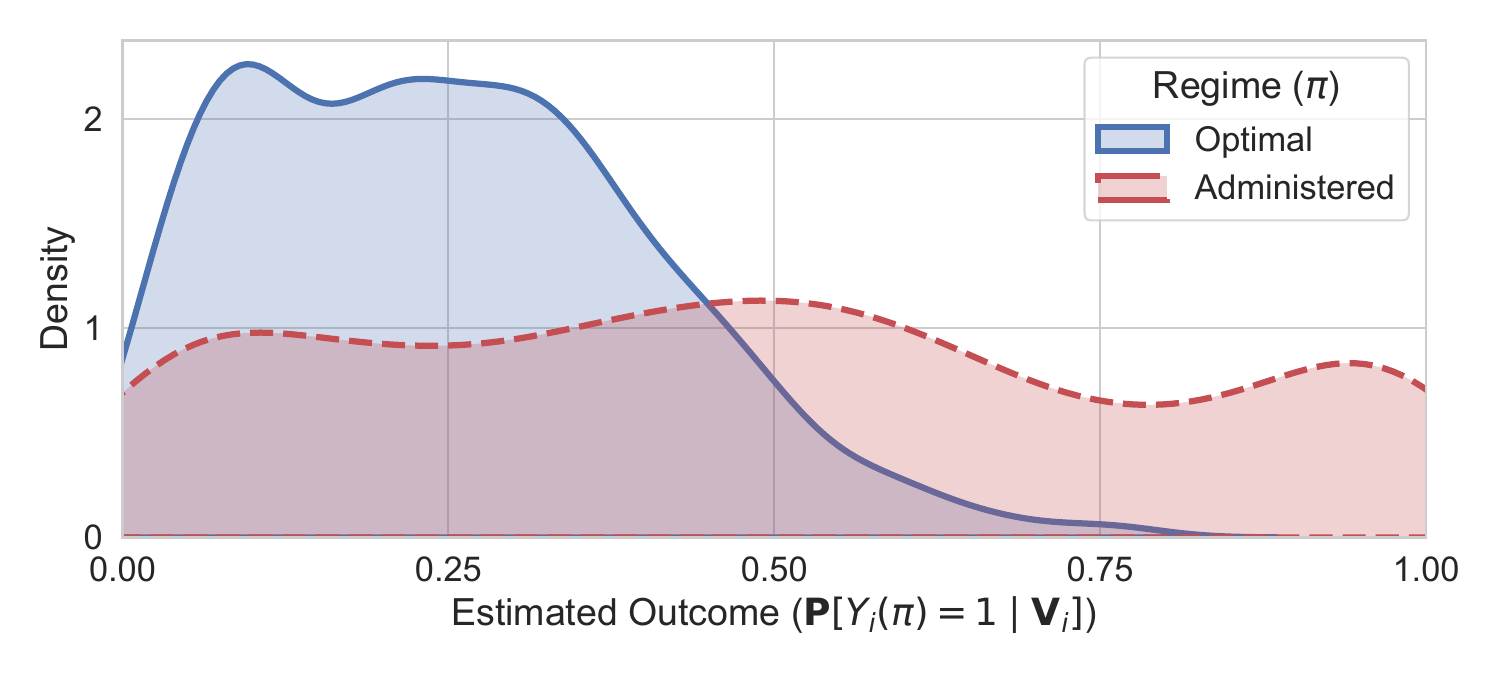} & \includegraphics[width=0.39\linewidth]{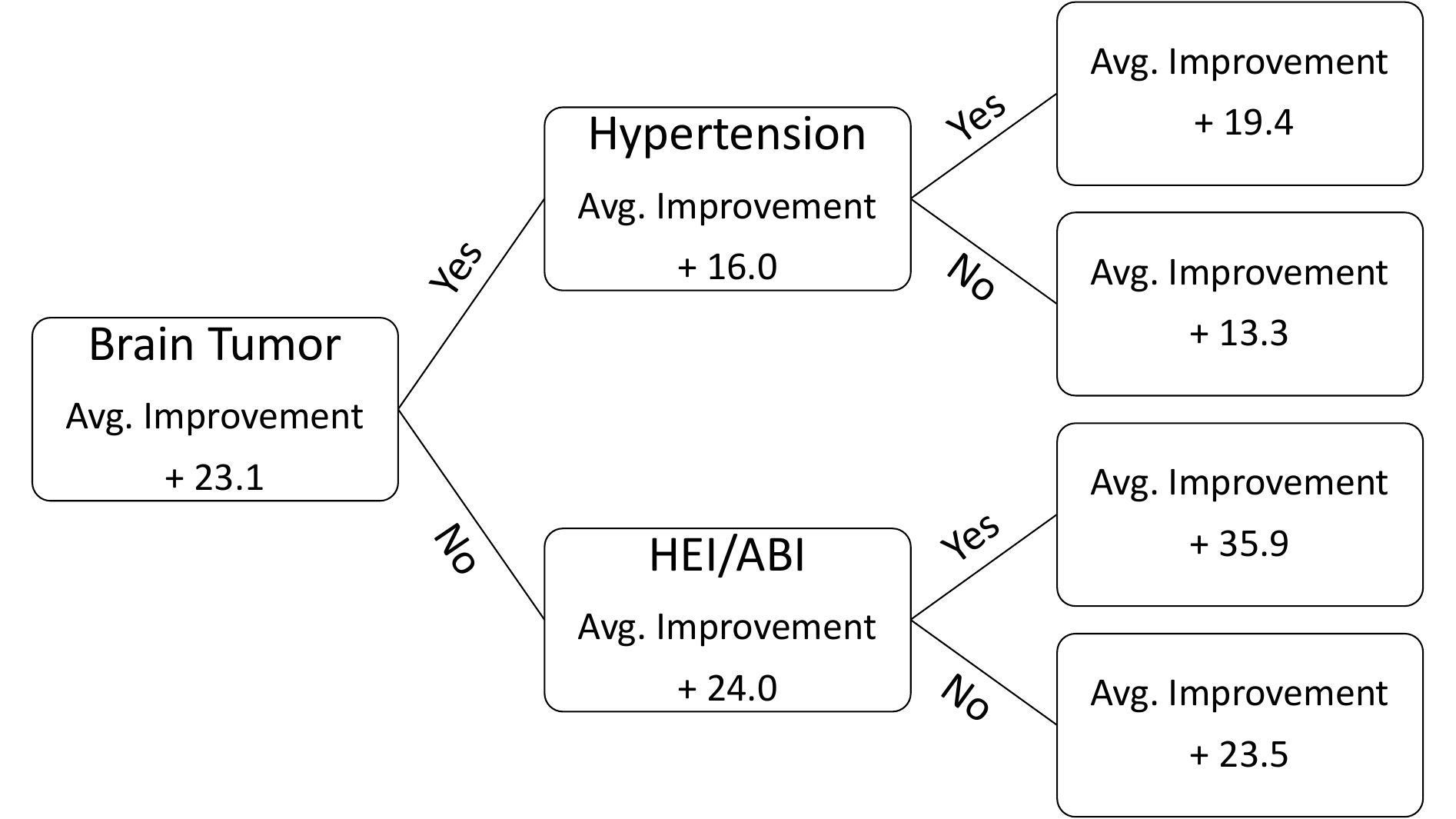}\\(a) & (b) 
    \end{tabular}
    \caption{(a) Estimated density of the outcome probabilities under optimal and clinician's administered policies. (b) Tree characterizing the subpopulations that would have benefited the most by switching to the optimal policy. The value at each node in the tree shows the percentage point \textit{improvement} in the outcome. Here, HEI/ABI refers to hypoxic-ischemic encephalopathy (HIE) and anoxic brain injury (ABI).}
    \label{fig:outcome}
\end{figure*}
\begin{figure*}
    \centering
    \begin{tabular}{ccc}
         \includegraphics[width=0.31\linewidth]{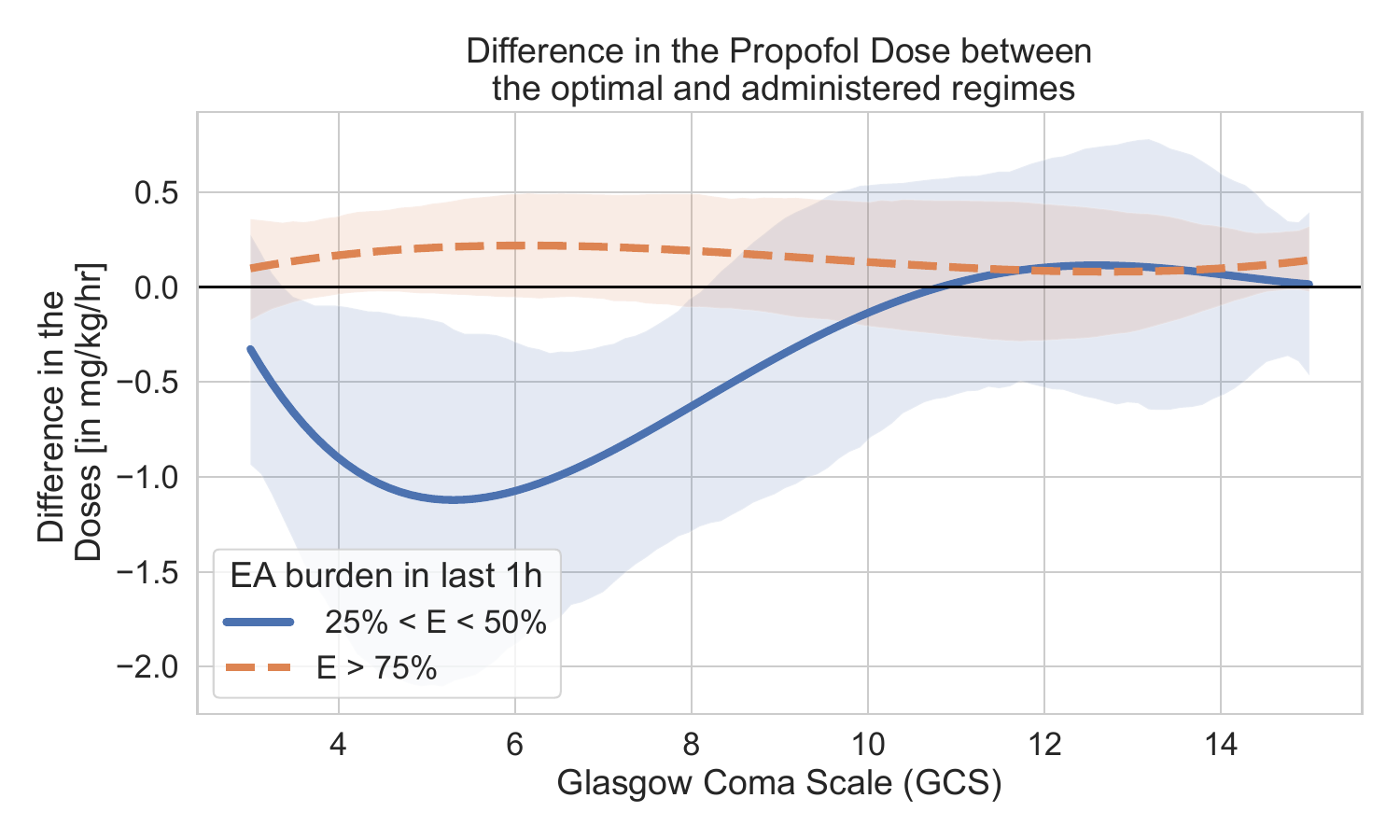} &
         \includegraphics[width=0.31\linewidth]{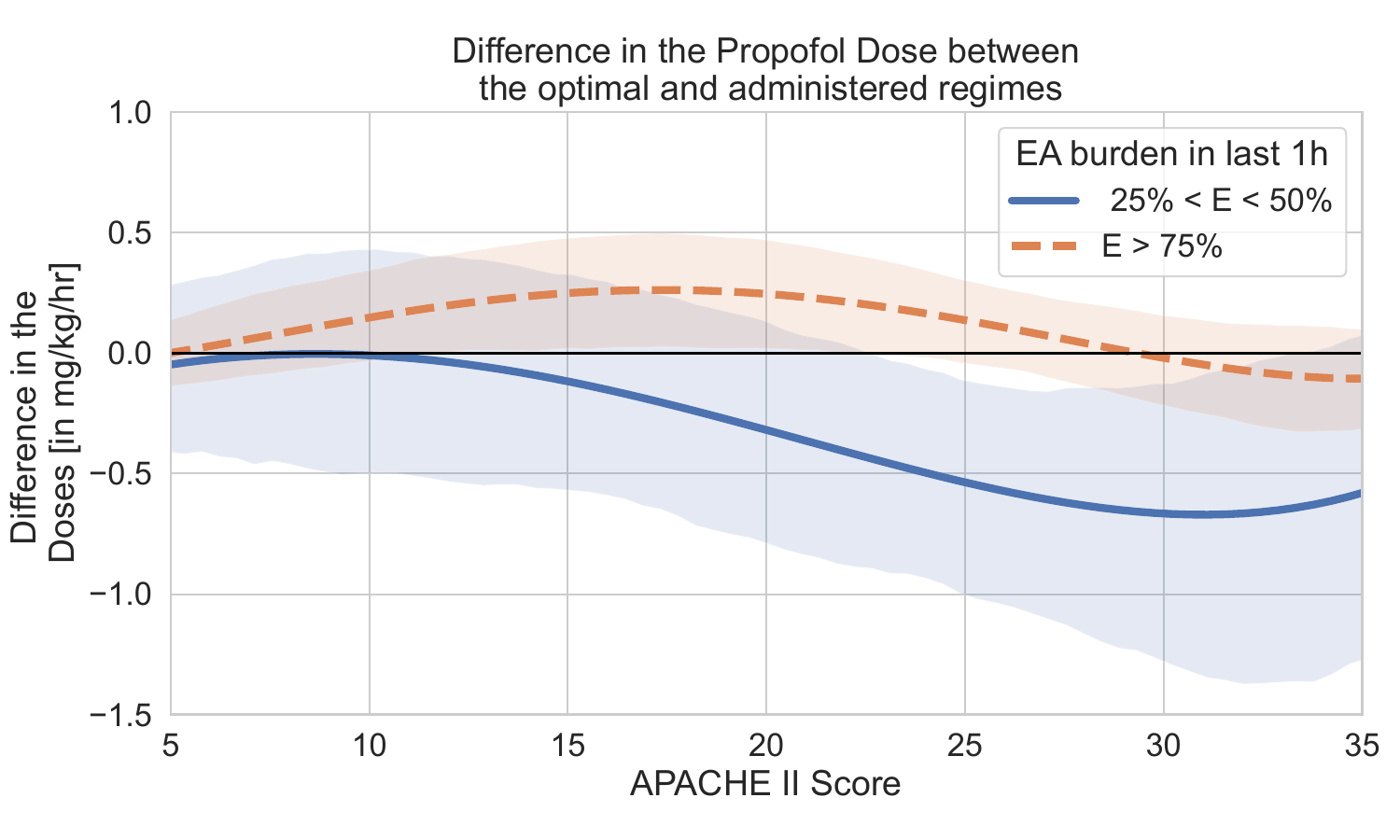} &
         \includegraphics[width=0.31\linewidth]{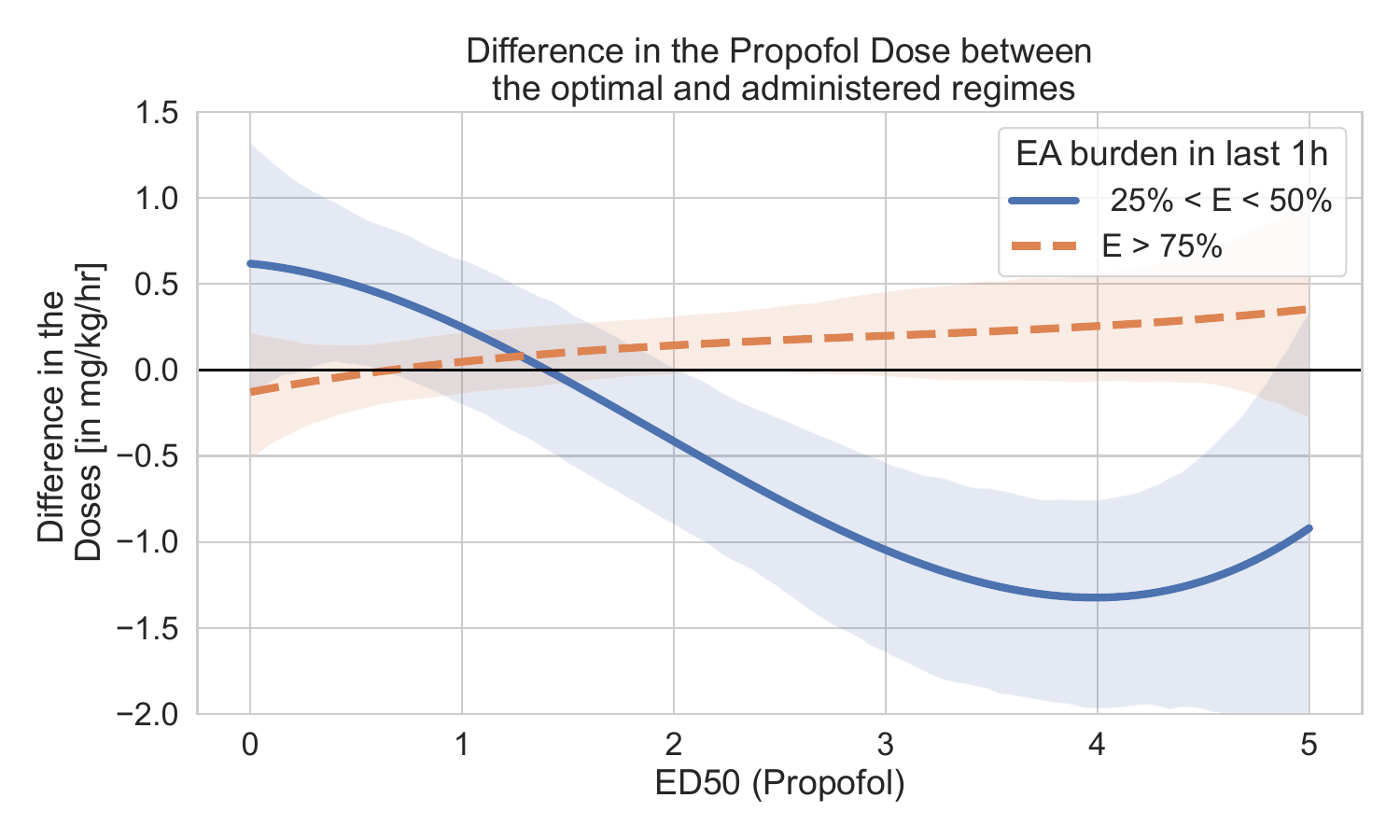} 
         \\
         (a) & 
         (b) & 
         (c) 
    \end{tabular}
    \caption{Difference in the propofol drug doses between the optimal and the administered regimes for mild and severe EA burden in last 1h for (a) patients on various levels of Glasgow coma scale (GCS); (b) patients with various levels of APACHE II scores; and (c) patients with various levels of ED50 for propofol, an important pharmacodynamic parameter determining the amount of drug required to reduce EA burden by 50\%.}
    \label{fig: prognoses}
\end{figure*}
\paragraph{Comparison Baselines.} We compare our approach to 49 approaches based on 10 different state-of-the-art finite timestep backward induction, infinite horizon, and deep reinforcement learning frameworks. The vast majority of methods cannot be run on our data setup out of the box and often require major modifications. The various approaches we compare use different underlying models, ways to discretize continuous outcomes, and predefined reward functions. We outline the methods we compare to and the implementation details in Appendix~\ref{sec: appendix_comp_method}.

\paragraph{Data Generation Procedure.} Our data-generative procedure is designed to emulate the real-world scenario where critically ill patients undergo drug treatment that affects their state. We design the data generation process to be customizable in five important aspects to discern how various methods perform with the challenges present in our real-world data: (i) number of covariates; (ii) number of total timesteps $\tau_i$, for each patient; (iii) number of unobserved timesteps, $\tau_i - T_i$, for each patient; (iv) cardinality of the action space; and (v) observed policies. We construct a total of 32 different experimental setups by varying these aspects. We provide the full details of our data generation process and experimental setups in Appendix~\ref{sec: dgp}. 

\paragraph{Results.} For a real-world data simulation, we use 1000 simulated ``patients'' with (i) 100 pre-treatment covariates, (ii) varying lengths of stay (10-15 timesteps), and (iii) unobserved timesteps (2-5 steps), where (iv) drug doses at each timestep are between 0 to 100 and (v) determined using an educated policy akin to one doctors use in the ICU. We display the percent of patients with poor outcomes under the proposed policies of our method, representative approaches from each of our comparison baseline categories, and predetermined approaches like inaction, random assignment, and max-dose in the left plot of Figure~\ref{fig: synthetic-sim1}. The right plot of Figure~\ref{fig: synthetic-sim1} shows results for the same setting except with (iii) no missing timesteps. In each of these complex setups, our matching-based method consistently yields optimal treatment policies, surpassing all comparison methods. Notably, among the 8 setups with 10-15 total timesteps and observed data generated from an educated policy, our method is consistently the top performer. 

\paragraph{Analysis.} Existing methods falter on simulated data emulating our real-world setup for various reasons. The suboptimal performance of Q-learning is likely caused by its inability to handle missing states as well as continuous action spaces \citep{huang2022reinforcement}. Infinite horizon methods like fitted Q-iteration mainly rely on a predefined reward function, often focusing on short-term objectives, and cannot handle continuous action spaces \citep{clifton2020q}. Deep RL methods like DDPG are also likely struggling with having to rely on a predefined reward function and the relatively small dataset size \citep{riachi2021challenges, kondrup2023towards, kang2023beyond, kalweit2017uncertainty}. More modern Deep RL methods like CQL, CRR, and BCQ mediate the deficiencies of DDPG. However, unlike our approach, these methods are inherently uninterpretable and, therefore, are unsuitable for high-stakes problems. 

Other methods can perform as good or better than our method when aspects of the data generating process are varied to look less like our real-world data. For example, approaches like infinite horizon and some Deep RL methods perform better when the observed data is generated from a random, rather than an educated, policy and backward induction methods perform better when there are fewer and no unobserved timesteps.

In Appendix~\ref{sec: appendix_synth_exp}, we thoroughly compare our method to the 49 baselines using 32 simulation setups. These results underscore the suboptimal performance of existing methods in scenarios with missing data, continuous action space, and highly stochastic state dynamics. Our method can handle these various challenges, allowing it to accurately estimate interpretable optimal regimes that are safe for high-stakes settings.


\section{TREATING SEIZURES IN CRITICALLY ILL PATIENTS}\label{sec: seizure_study}
We now present the analysis and insights derived from our optimal treatment estimation approach when applied to a cohort of 995 critically ill patients. This cohort is comprised of individuals aged 18 and older with confirmed electrographic EA as diagnosed by clinical neurophysiologists or epileptologists.

We evaluate our approach by comparing the estimated optimal treatment policy $P( Y_i(\pi^*_i)=1 | \V_i)$ with the clinician's administered policy $P( Y_i(\pi_i)=1 | \V_i)$ for each patient. Our analysis indicates a significant improvement in patient outcomes, with a 23.6 $\pm$ 1.9 percentage point reduction in the probability of adverse events under the optimal regimen. Few patients under the optimal policy had over a 50\% chance of an adverse outcome (Figure~\ref{fig:outcome}(a)). Figure~\ref{fig:outcome}(b) reveals that patients with hypoxic-ischemic encephalopathy (HIE) or anoxic brain injury (ABI) experienced a substantial 35.9 percentage point decrease in the likelihood of an adverse outcome, highlighting those who benefited most from our estimated optimal treatment policies.

\begin{table}
    \centering
\caption{APACHE II scores and corresponding non-operative mortality or death rate from \cite{knaus1986apache}, as well as estimated $Y$ under estimated administered regime and optimal regime.}
\label{tab:apache_table}
\resizebox{\linewidth}{!}{
    \begin{tabular}{l|c|cc}
    \hline
        \begin{tabular}{l}
         APACHE\\II Score
         \end{tabular} &  
         \begin{tabular}{c}
         Death\\ Rate
         \end{tabular} & 
         \begin{tabular}{c}
         Est. \\$\E[Y_i(\pi_i)]$
         \end{tabular} & 
         \begin{tabular}{c}
        Est. \\$\E[Y_i(\widehat{\pi}^*_i)]$
         \end{tabular} 
         \\
         \hline
         0 to 4&  4\%&  17\%& 6\%\\
         5 to 9&  8\%&  22\% & 8\%\\
         10 to 14&  15\%&  35\% & 17\%\\
         15 to 19&  24\%&  48\%& 25\%\\
         20 to 24&  40\%&  56\%& 31\%\\
         25 to 29&  55\%&  61\%& 35\%\\
         30 to 34&  73\%&  73\%& 36\%\\\hline
    \end{tabular}}

\end{table}

 We compare and contrast the optimal regimes with the administered regimes for each drug. We consider the variability of each drug's regime with respect to patients' pre-treatment prognosis measured as APACHE II score \citep{knaus1986apache} and Glasgow coma scale (GCS) \citep{jain2018glasgow}. APACHE II score quantifies disease severity in ICU patients and GCS measures impaired consciousness in acute medical and trauma patients. Both of these measures are clinically relevant for deciding treatment strategies \citep{mumtaz2023apache}. Table~\ref{tab:apache_table} displays mortality rates from \citet{knaus1986apache} and estimated $Y$ under administered and optimal regimes for different APACHE II scores. The optimal regime improves outcomes across all levels, with the most benefits seen in patients with high APACHE II scores (i.e., with worse prognoses).

\textbf{Propofol Regimes.} Figures~\ref{fig: prognoses}(a)~and~\ref{fig: prognoses}(b) show that, on average, the estimated optimal propofol dose for individuals with low EA burden is generally lower than the administered dose, especially for those with worse prognoses (lower GCS or higher APACHE II scores). Conversely, when patients have a severe EA burden in the last hour and an APACHE II score below 30, the optimal dose is marginally higher than the administered dose. Also, one must adjust propofol dosages based on patients' PK/PD, specifically, based on the ED50 values -- a PD parameter quantifying the amount of drug required to reduce the EA burden by 50\%. When the EA burden is low, we recommend increasing the dosage for patients with low ED50 values to alleviate EA and decreasing it for those with high ED50 values, as an excess of propofol may lead to adverse effects (see Figure~\ref{fig: prognoses}(c)).

\begin{figure}
    \centering
        \begin{tabular}{cc}
        \includegraphics[width=0.78\linewidth]{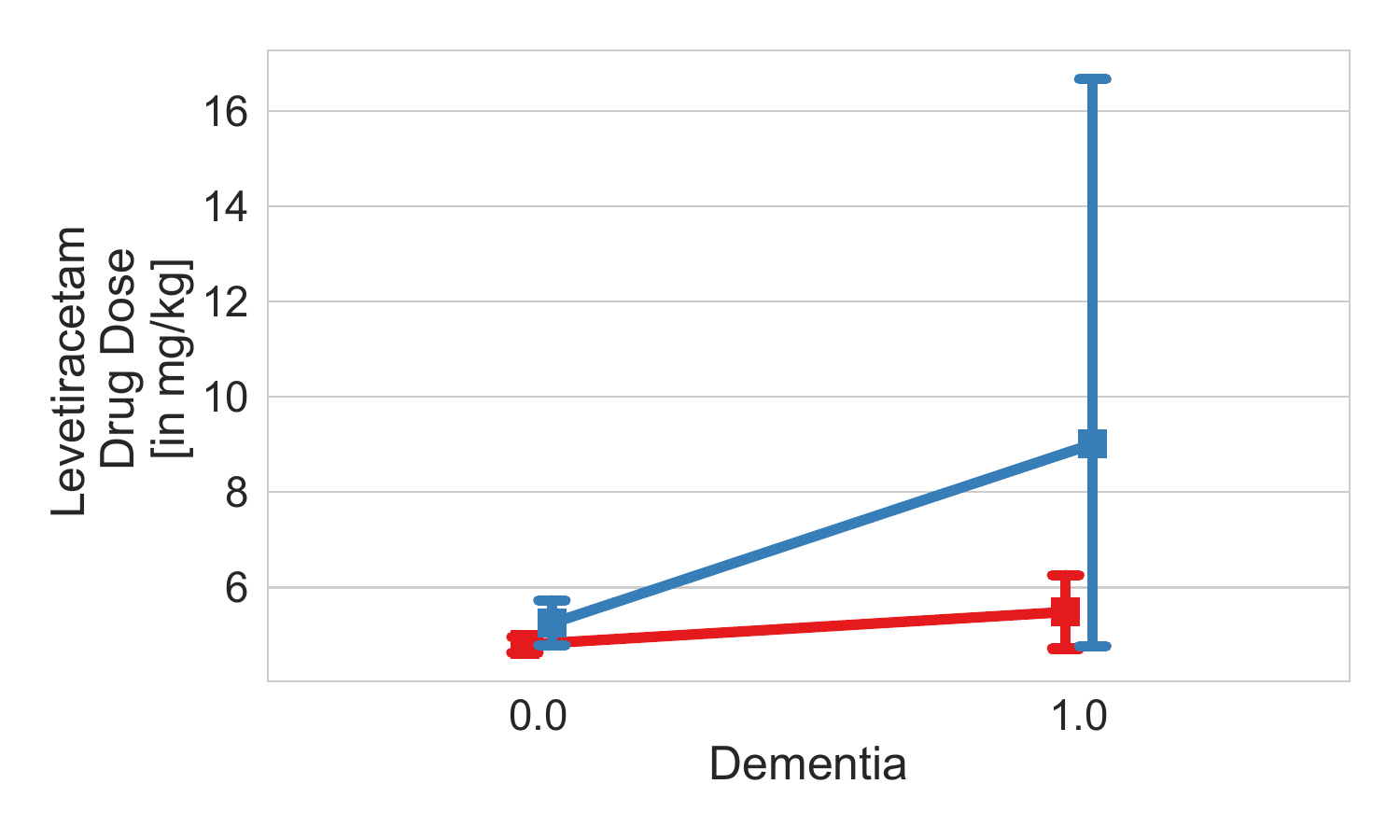} 
         \\
         (a)\\
         \includegraphics[width=0.78\linewidth]{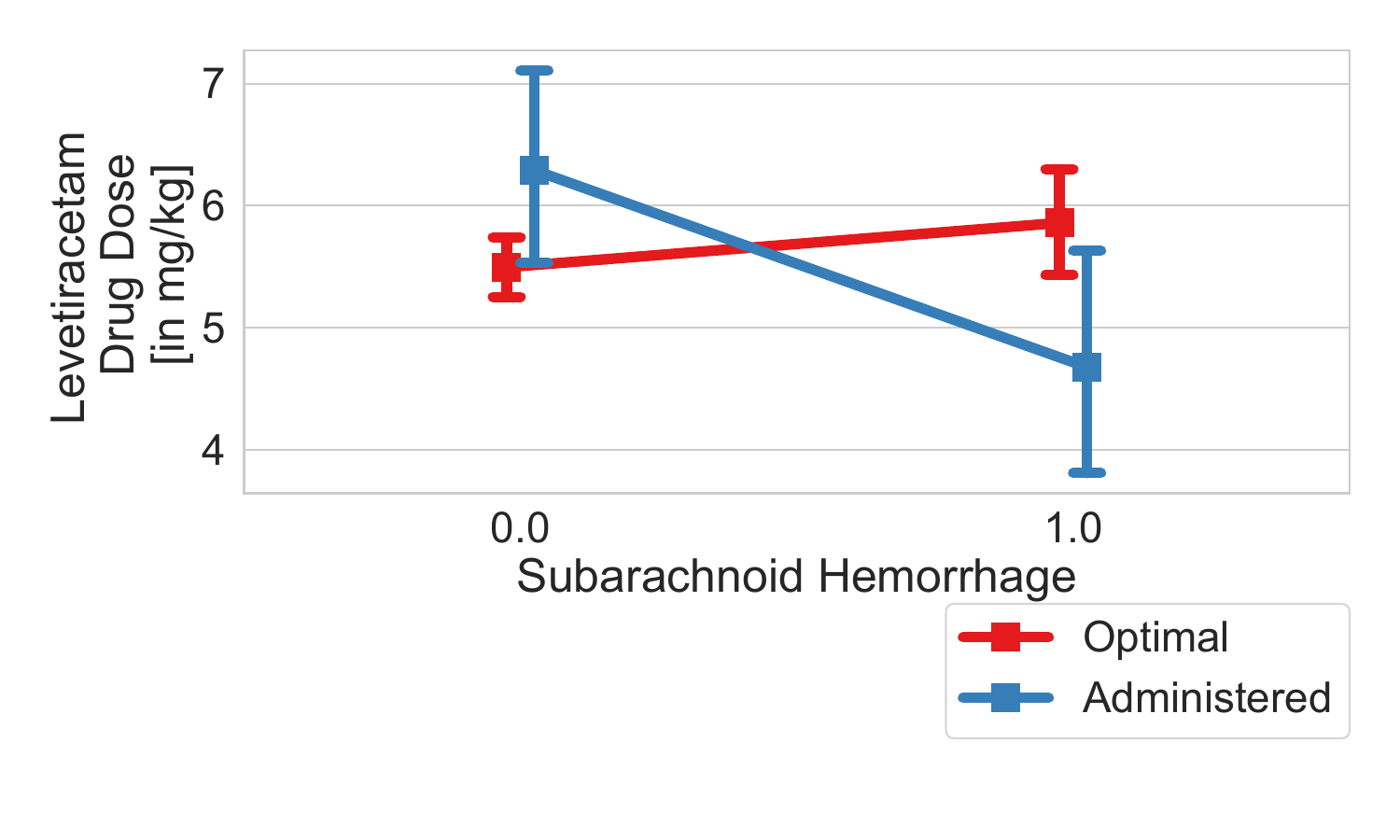} \\
         (b) 
    \end{tabular}
    \caption{Difference in the levetiracetam doses between the optimal and the administered regimes for (a) patients with and without dementia experiencing a sustained EA burden for 12 hours; and (b) patients with and without subarachnoid hemorrhage experiencing a sustained EA burden for 6 hours.}
    \label{fig: lev_diff}
\end{figure}

\textbf{Levetiracetam Regimes.} 
The optimal and administered levetiracetam regimes generally align, except for patients with sustained 12-hour EA burden. In such cases, the optimal regime recommends a lower dose (0.50 mg/kg on average) compared to the administered regime (0.82 mg/kg on average). For dementia patients, the difference is more pronounced, with the optimal regime suggesting a dose of 4.2 mg/kg lower (see Figure~\ref{fig: lev_diff}(a)). Conversely, subarachnoid hemorrhage patients with a 6-hour sustained EA burden receive a 1 mg/kg higher dose with the optimal regime (see Figure~\ref{fig: lev_diff}(b)).


To summarize, our findings indicate that patients in this study would, on average, be less likely to have an adverse outcome under the optimal regimes estimated by using our method. These optimal regimes would lead us to advocate for an assertive approach to managing the high EA burden in more critically ill patients while reducing propofol and levetiracetam dosages for relatively healthier patients or those with mild EA.
\section{DISCUSSION \& CONCLUSION}
We present an approach that is capable of handling many challenges with real-world observational data like variable timesteps, missing states, a continuous action space, and small data size. Our approach balances accuracy and interpretability and demonstrates superior performance through simulation. We ultimately operationalize our approach to learn treatment regimes for ICU patients with EA, showcasing its ability to solve real-world problems.

\paragraph{Clinical Relevance.} 
The current absence of evidence-based guidelines to inform ASM regimes (drug type and dosing) in patients with EA results in frequent overprescription of ASMs in response to EA \citep{zafar2020electrographic, rubinos2018ictal}. High EA-burden is frequently treated with escalating doses of ASMs and anesthetics, and many
of these patients are also discharged on ASM treatment \citep{zafar2018effect, tabaeizadeh2020burden, dhakar2022epileptiform, alvarez2017use, kilbride2009seizure, punia2020post}. Our findings suggest that not all
patients may benefit from such ASM escalation. Thus, careful consideration of the baseline illness severity, injury type, and patient comorbidities is important to determine the risk-benefit trade-off of initiating treatment and selecting treatment intensity. For example, patients with cognitive impairment and dementia have a higher risk of ASM adverse effects \citep{mendez2003seizures, cretin2021treatment} and may require lower-intensity treatment, which is supported by our findings. Finally, as shown in Figure~\ref{fig: prognoses} and Figure~\ref{fig: lev_diff}, heterogeneous treatment responses need to be considered in selecting drug dosing. Current clinical practice relies on population-level pharmacological data to infer standardized dosing regimens used for all patients. 
However, this one-size-fits-all approach is suboptimal due to the patient-level PK/PD heterogeneity shown in our study (see Figure~\ref{fig: prognoses}(c)).
Our findings strongly support the need for \textit{clinical trials} to reveal heterogeneous causal effects and construct individualized optimal treatment. Such efforts can guide evidence-based clinical practice and improve patient care in the ICU. 

\textbf{Limitations.} Like all causal research, our study relies on untestable assumptions. We assume there are no hidden variables affecting both EA burden and patient discharge outcomes, though unmeasured disease characteristics might violate this assumption. Additionally, the misspecification of our predefined policy template, intended for doctor interpretability, could affect real-world drug administration, akin to issues discussed in recent work \cite{savje2023causal}. Furthermore, while we focus on point estimation for personalized optimal treatment regimes, handling uncertainty, especially when estimating the exposure map from observed data, remains an open question.

\textbf{Future Direction.} Addressing the limitations inherent in our approach, we identify three promising areas for future work. First, there is a need for research into uncertainty quantification for estimated personalized optimal treatment regimes, with broader implications for situations where exposure mapping is data-driven. Second, developing a non-parametric approach for sensitivity analysis and partial identification has the potential to advance research in this area. Third, interpretable Deep RL has been explored to a limited capacity in works like \cite{lyu2019sdrl} and \cite{li2023differentiable}. Given the relatively strong performance of these methods, future work that optimizes Deep RL for offline and off-policy tasks is a promising direction for future work in this area.
\section*{Acknowledgments}
We acknowledge funding from the National Science Foundation and Amazon under grant NSF IIS-2147061, and the National Institute on Drug Abuse under grant DA056407. National Institute on Drug Abuse R01DA056407 supports Harsh Parikh. NSF grant DMS-2046880 supports Cynthia Rudin, Alexander Volfovsky, and Quinn Lanners. Alexander Volfovsky is also supported by a National Science Foundation Faculty Early Career Development Award (CAREER: Design and analysis of experiments for complex social processes). The authors want to thank Dr. Lina Montoya and Srikar Katta for their insightful and constructive comments.
\bibliographystyle{apalike}
\bibliography{biblio}
\section*{Checklist}

 \begin{enumerate}

 \item For all models and algorithms presented, check if you include:
 \begin{enumerate}
   \item A clear description of the mathematical setting, assumptions, algorithm, and/or model. [\framebox{Yes}/No/Not Applicable]. Section~\ref{sec: prelim} and Section~\ref{sec: id} discuss the mathematical settings and the relevant assumptions. Section~\ref{sec: method} discusses our methodology.
   \item An analysis of the properties and complexity (time, space, sample size) of any algorithm. [Yes/No/\framebox{Not Applicable}]
   \item (Optional) Anonymized source code, with specification of all dependencies, including external libraries. [\framebox{Yes}/No/Not Applicable]. \href{https://github.com/almost-matching-exactly/opt\_tx\_regime\_matching}{GitHub Link} (https://github.com/almost-matching-exactly/opt\_tx\_regime\_matching)
 \end{enumerate}

 \item For any theoretical claim, check if you include:
 \begin{enumerate}
   \item Statements of the full set of assumptions of all theoretical results. [\framebox{Yes}/No/Not Applicable]. Section~\ref{sec: prelim}, Section~\ref{sec: id} and Section~\ref{sec: method} state the relevant assumptions for our Proposition 1.
   \item Complete proofs of all theoretical results. [\framebox{Yes}/No/Not Applicable]. Appendix~\ref{sec: appendix_theorem} provides the proof of the proposition.
   \item Clear explanations of any assumptions. [\framebox{Yes}/No/Not Applicable]. Section~\ref{sec: prelim}, Section~\ref{sec: id}, and Section~\ref{sec: method} discuss the relevant assumptions for our Proposition 1.
 \end{enumerate}

 \item For all figures and tables that present empirical results, check if you include:
 \begin{enumerate}
   \item The code, data, and instructions needed to reproduce the main experimental results (either in the supplemental material or as a URL). [\framebox{Yes}/No/Not Applicable]. \href{https://github.com/almost-matching-exactly/opt_tx_regime_matching}{Click here to access our GitHub containing the code.} The patients' data will be provided to Dr M Brandon Westover, and Dr Sahar F Zafar on appropriate request.
   \item All the training details (e.g., data splits, hyperparameters, how they were chosen). [\framebox{Yes}/No/Not Applicable] Appendix~\ref{sec: appendix_comp_method} and Appendix~\ref{sec: appendix-add-synth-details} discuss training details. Code also includes comments.
    \item A clear definition of the specific measure or statistics and error bars (e.g., with respect to the random seed after running experiments multiple times). [\framebox{Yes}/No/Not Applicable]
    Contents of each plot is described in the corresponding caption. Further discussion of experimental details is in Appendix~\ref{sec: appendix_comp_method}.
    \item A description of the computing infrastructure used. (e.g., type of GPUs, internal cluster, or cloud provider). [\framebox{Yes}/No/Not Applicable]
    Discussed in Appendix~\ref{sec: appendix-add-synth-details}
 \end{enumerate}

 \item If you are using existing assets (e.g., code, data, models) or curating/releasing new assets, check if you include:
 \begin{enumerate}
   \item Citations of the creator If your work uses existing assets. [\framebox{Yes}/No/Not Applicable] Citations to existing assets are in Appendix~\ref{sec: appendix_comp_method}.
   \item The license information of the assets, if applicable. [Yes/No/\framebox{Not Applicable}]
   \item New assets either in the supplemental material or as a URL, if applicable. [Yes/No/\framebox{Not Applicable}]
   \item Information about consent from data providers/curators. [\framebox{Yes}/No/Not Applicable]
   \item Discussion of sensible content if applicable, e.g., personally identifiable information or offensive content. [\framebox{Yes}/No/Not Applicable]
 \end{enumerate}

 \item If you used crowdsourcing or conducted research with human subjects, check if you include:
 \begin{enumerate}
   \item The full text of instructions given to participants and screenshots. [Yes/No/\framebox{Not Applicable}]
   \item Descriptions of potential participant risks, with links to Institutional Review Board (IRB) approvals if applicable. [\framebox{Yes}/No/Not Applicable] We will add the IRB approvals from involved institution with the camera-ready version (post-acceptance)
   \item The estimated hourly wage paid to participants and the total amount spent on participant compensation. [Yes/No/\framebox{Not Applicable}]
 \end{enumerate}

 \end{enumerate}

\begin{appendices}
    \newpage
\onecolumn
\aistatstitle{Safe and Interpretable Estimation of Optimal Treatment Regimes: Supplementary Materials}
\section{\textsc{Dynamic Treatment Regime \& Reinforcement Learning Literature Survey}}\label{sec: lit_survey_appendix}
There are a number of different techniques for estimating optimal treatment regimes. Prior methods include parametric, semi-parametric, and non-parametric modeling approaches and are often combined with reinforcement learning (RL) frameworks such as Q-learning and policy gradient. We categorize the existing methods into five categories: \textit{Finite Timestep Backward Induction}, \textit{Infinite Time Horizon}, \textit{Censored Data}, \textit{Deep Reinforcement Learning}, and \textit{Causal Nearest Neighbors}. Methods from each of these categories excel in certain settings. However, in this section, we highlight the limitations of each approach that ultimately make them unsuitable for our complex, high-stakes problem.

Finite timestep backward induction methods make up the majority of optimal treatment policy estimation methods. \cite{murphy2003optimal} and \cite{robins2004optimal} were some of the first ones to utilize backward induction in a semiparametric approach using approximate dynamic programming. \cite{murphy2005generalization} introduced the now widely used Q-learning, of which initial extensions focused on using parametric and semi-parametric modeling of the Q-functions \citep{moodie2010estimating, chakraborty2010inference, song2015penalized}. These approaches can produce interpretable policies and can be easier to implement. However, correct specification of the Q-functions can be difficult, particularly with observational data \citep{moodie2014q}. This can lead to poor estimates of the optimal policy when a misspecified linear model is used. For this reason, recent work has focused on using flexible non-parametric machine learning methods \citep{zhang2012estimating, zhao2015new, murray2018bayesian, blumlein2022learning}, particularly within the Q-learning framework \citep{qian2011performance, moodie2014q, zhang2018interpretable}. While these methods are less prone to model misspecification, they often result in complex treatment regimes for which the rationale behind the treatment decision is difficult to discern. Although, \cite{blumlein2022learning} and \cite{zhang2018interpretable} proposed more explainable nonparametric approaches. 

The majority of backward induction methods assume all patients have the same number of fixed timesteps, which presents difficulty when working with variable timesteps across patients and unobserved states. Infinite horizon methods, like fitted Q-iteration \citep{ernst2005tree, clifton2020q} and  \cite{ertefaie2018constructing}'s Q-learning approach, are better suited to handle these complexities. However, these methods necessitate a reward value to be associated with every action taken by each unit. These reward values are often assumed to be a measurable value that is intrinsically linked to the optimization problem. When this is not the case, they need to be calculated using a predefined function over the observed variables. Having to create such a reward function is often a difficult task that can lead to poor optimal regime estimates \citep{mataric1994reward, koenig1996effect, singh2010intrinsically}. Other work has investigated using backward induction with censored data \citep{goldberg2012q, lyu2023imputation, zhao2020constructing}. However, these methods have focused on survival analysis time-to-event tasks, which differ from our setup where we have a labeled outcome for each patient.

Regardless of time-step constraints, all of the methods discussed thus far assume that there are a discrete number of treatment options at each time point. Furthermore, while there is extensive work on backward induction methods for observational data \citep{moodie2012q}, many methods impose a strong positivity assumption over all of the treatments at each timepoint \citep{qian2011performance, zhao2015new, blumlein2022learning}. This assumption is often broken in observational data. For example, in the medical setting patient care is given under the supervision of a trained professional and thus, unless randomized, at any given time point a patient in a particular state may have a near-zero chance of receiving a particular treatment. While approaches like \cite{schulte2014q} do employ weaker positivity assumptions, there is limited discussion on how various backward induction methods handle extremal propensity scores.

Deep reinforcement learning (RL) methods are a fast-growing area of research for optimal treatment regime estimation. Deep RL methods can be categorized as online or offline and on-policy or off-policy. In real-world high-stakes settings online and on-policy methods are infeasible, limiting the scope of applicable methods to offline, off-policy approaches. \cite{mnih2013playing} introduced Deep Q-Learning as an effective method for off-policy RL and \cite{lillicrap2015continuous} extended this method to a continuous action space with deep deterministic policy gradient (DDPG). More recent work has focused on improving upon DDPG by improving sampling efficiency \citep{haarnoja2018soft}, limiting overestimation bias \citep{fujimoto2018addressing, kumar2020conservative}, overcoming extrapolation error \citep{fujimoto2018off}, and using a critic-regularized approach \citep{wang2020critic}. 

Deep RL methods are capable of learning complex optimal treatment regimes and can handle variable and infinite timesteps. These methods are significantly more data and resource-hungry than non-deep learning approaches. Although, Deep RL methods like \cite{tschantz2019scaling} and \cite{haarnoja2018soft} offer improvements in this area. A larger issue with Deep RL is that it requires reward values to be associated with each action. This can cause issues similar to those discussed with infinite horizon methods. A possible solution to not having reward values for infinite horizon and Deep RL methods is to use inverse reinforcement learning to learn a good reward function \citep{ng2000algorithms, arora2021survey}. However, such an approach would add an additional layer of complexity to the estimation procedure. In the case of Deep RL, this would further exacerbate what is already its most crucial limitation in its inherent lack of interpretability. The black-box nature of Deep RL makes it a poor choice for optimal treatment regime estimation in high-stakes applications. 

The previously mentioned methods either assume (i) the correct specification of a reward function or (ii) that there are no missing states or actions leading up to the final outcome. These assumptions do not align with our real-world scenario.

Matching is an intuitive method for optimal treatment regime estimation. Despite its inherent interpretability, little work has been done in this area. \cite{zhou2017causal} used a nearest-neighbor approach that examined the causal treatment effects within neighborhoods of similar patients to estimate optimal treatment regimes. While mentioning that their method can be extended to observational studies, they focus on randomized controlled trials - lacking theoretical or experimental results for the observational setting. Furthermore, they only consider a singular timestep with discrete treatment options and use a limited univariate approach for matching in high dimensions. Ultimately, their matching approach shows promise as an accurate and interpretable approach to optimal treatment regime estimation but is unable to handle the complexities commonly found in real-world problems.

Ideally, we want a method that can handle continuous action and state spaces, missing timesteps, does not require a reward function to be specified, and can be trained on a small number of samples. Furthermore, we want a method that is interpretable given the high-stakes setting. Table~\ref{tab: lit-review}, in the main text, summarizes the different optimal treatment regime estimation approaches in regard to these desired attributes. In Section~\ref{sec: method}, we present our matching approach for optimal treatment regime estimation. We subsequently present results showing our method's superior performance over a number of comparison approaches across various settings (see Section~\ref{sec: synth} and Appendix~\ref{sec: appendix-add-synth-results}).  Ultimately, to the best of our knowledge, our method is the only approach that possesses all of the qualities needed to effectively address our problem.

\section{\textsc{Distance Metric Learning and Almost Exact Matching}}\label{sec: appendix_matching}
In this section, we discuss some recent and relevant work in the almost exactly matching and distance metric learning literature. In an ideal scenario, we would achieve exact matches for some units. However, in high-dimensional contexts with continuous covariates, exact matches are rare. When performing nearly exact matching with a caliper of $r$, the objective is to achieve a close match on relevant features while not being overly concerned about matching on irrelevant ones. Therefore, especially in cases with limited data, the choice of the distance metric $d$ for matching becomes crucial. Recent matching approaches have focused on distance metric learning before the matching process. One such approach, Genetic Matching \citep{diamond2013genetic}, employs a genetic algorithm to learn an appropriate distance metric. However, it has been found to perform poorly for individualized estimation and is limited to binary or categorical exposures. Another method, Matching After Learning to Stretch (MALTS) \citep{parikh2022malts}, is effective for individualistic estimation but struggles to converge in high-dimensional settings with small datasets. A recent approach called Variable Importance Matching (VIM) \citep{lanners2023variable} uses a highly regularized model like LASSO or a shallow decision tree to model $\E[Y | \V ]$. It then utilizes the variable importance scores from the fitted model to guide the selection of the distance metric. This approach is both fast and interpretable and works well in high-dimensional scenarios, making it well-suited for our problem.
\section{\textsc{Pharmacokinetics and Pharmacodynamics}}\label{sec: appendix_pkpd}
In this section, we discuss our modeling choice for PK and PD mechanistic models. 
\paragraph{Pharmacokinetics.} We use a one-compartment PK model to estimate the concentration of drug $j$ for patient $i$ at time $t$ (${D}_{j,i,t}$) as:
\begin{equation}
    g_i(\{\D_{i,t'}\}_{t'=1}^{t-1}, \Z_{i,t}) = e^{-\gamma_{j,i}} {D}_{j,i,t-1} + Z_{j,i,t},
\end{equation}
where pharmacokinetic parameter $\gamma_{j,i}$ is proportional to the half-life of the drug $j$ in patient $i$. 

\paragraph{Pharmacodynamics.} We model PD using Hill's model \citep{nelson2008lehninger} to estimate the short-term effectiveness of the ASMs in reducing EA burden:
{\footnotesize
\begin{equation}\label{eq: pk-pb_burden}
    f_i(\{E_{i,t'}\}_{t'=1}^{t-1}, \D_{i,t}) = \beta_i\left(1 - \sum_j \frac{{D}_{j,i,t}^{\alpha_{j,i}}}{{D}_{j,i,t}^{\alpha_{j,i}}+ED50_{j,i}^{\alpha_{j,i}}}\right),
\end{equation}}

where $\beta_i$ is patient $i$'s EA burden when no drugs are administered, $\alpha_{j, i}$ models the affinity of drug $j$'s ligand to a receptor for patient $i$, and $ED50_{j, i}$ is the amount of drug concentration necessary to reduce EA burden by 50\% from the maximum level. 
\section{\textsc{Data Generative Mechanism for the Simulation Study}}\label{sec: dgp}
We base our synthetic data experiments on our real world application where patients experiencing seizures are treated with anti-seizure medications. For our synthetic experiments, we let the first-order pharmacological state-transition model outlined in Appendix~\ref{sec: appendix_pkpd} be the true model for each patient's drug response and EA burden progression. 

For each patient $i\in\{1,\dots,n\}$, the PK/PD model is defined by the following parameters: $\beta_i$,  $\gamma_{i,j}$, $\alpha_{i,j}$, and $ED50_{i,j}$ for each drug $j$. For simplicity, and to allow for comparison to more methods, we consider a setting with only one drug. Associated with each patient are $p$ pre-treatment covariates, $X_{i,1},\dots, X_{i,p} \overset{iid}{\sim} \text{Normal}(0, 1)$. We let the PK/PD parameters be correlated with the pre-treatment covariates $\X_i$ such that $\beta_i \sim \text{Normal}\left(100 + 10X_{i,1}, 5 \right)$
and 
$ED50_i \sim \text{Normal}\left(15 - 2X_{i,3}, 1\right)$. 
Further, $\gamma_i, \alpha_i \overset{iid}{\sim} \text{Normal}(1, 0.1)$.

From here, we let the total number of timesteps, $\tau_i$, be a random integer in [${T}_{min}$, ${T}_{max}$] and set the number of observed states as $T_i = \tau_i - m_i$, where $m_i$ is the number of unobserved timesteps and is a random integer in [${M}_{min}$, ${M}_{max}$]. Finally, $E_{i,0}$, the initial burden for patient $i$, is sampled as $E_{i,0} \sim \text{Normal}(75 + 5X_{i,2}, 5)$, and is lower bounded by 0 and upper bounded by $\beta_i$.

We simulate a complete sequence of states $\{E_{i,t}\}_{t=1}^{\tau_i}$ and actions $\{Z_{i,t}\}_{t=1}^{\tau_i}$ given the initial burden $E_{i,0}$, a policy $\pi_{i}$, and the patient's corresponding PK/PD parameters. We use the same PK/PD equations outlined in Appendix~\ref{sec: appendix_pkpd} with a small amount of noise added to the patient's EA burden at each timestep. In particular, we calculate the EA burden for patient $i$ at timestep $t$ by slighltly modifying Equation~\ref{eq: pk-pb_burden} in Appendix~\ref{sec: appendix_pkpd} so that 
\begin{equation}
    E_{i,t} = \beta_i\left(1 - \sum_j \frac{{D}_{i,t}^{\alpha_{i}}}{{D}_{i,t}^{\alpha_{i}}+ED50_{i}^{\alpha_{i}}}\right) + \epsilon_{E_{i,t}}.
\end{equation}
where $\epsilon_{E_{i,t}} \sim \text{Normal}(0, 2.5)$. This produces a series of EA burdens $\{E_{i,t}\}_{t=1}^{\tau_i}$ drug doses $\{Z_{i,t}\}_{t=1}^{\tau_i}$ and drug concentrations $\{D_{i,t}\}_{t=1}^{\tau_i}$ corresponding to each patient $i$. The outcome is related to the patient's pre-treatment covariates, EA burdens, and drug concentrations - thus inducing a level of confounding. In particular, we calculate the continuous outcome value as

\begin{equation}\label{eq: out-func}
O_i = \frac{1}{\tau_i} \left[\exp{\left(\sum_{j=1}^{2} \frac{X_{i,j}}{2}\right)} \left( \sum_{t=1}^{\tau_i} \exp{\left(\frac{E_{i,t}}{50}\right)} - 1 \right) + \exp{\left(\sum_{j=3}^{4} \frac{X_{i,j}}{2}\right)} \left( \sum_{t=1}^{\tau_i} \exp{\left(\frac{D_{i,t}}{50}\right)} - 1 \right) \right]
\end{equation}

Note that we desire a smaller continuous outcome value. This outcome function represents a scenario where patients with a large average value in $X_{i,1}$ and $X_{i,2}$ are more at risk from high levels of EA burden. Whereas, patients with a large average value in $X_{i,3}$ and $X_{i,4}$ are more at risk from high drug concentrations. Finally, to emulate the real-world setting where we observe a binary outcome, we discretize the continuous outcomes to a binary outcome for each patient, setting $Y_i = \mathbf{1}\left[O_i > 3\right]$. 
\begin{remark}
    Three was chosen as our cutoff value for the binary outcomes to create a setting where about 50\% of patients experience a bad outcome (i.e. $Y = 1$). By using a static value, we could more easily compare the binary outcomes across a variety of data generation setups. 
\end{remark}

Ultimately, the observed data for each patient $i$ is $\{X_i, \{E_{i,t}\}_{t=1}^{T_i}, \{Z_{i,t}\}_{t=1}^{T_i}, Y_i\}$.
Note that the observed history only includes the states and actions up to timestep $T_i$, not $\tau_i$, and only includes the binary outcome $Y_i$, not $O_i$. 

\subsection{Data Generation Process Setups}\label{sec: appendix-dgp-setups}

We vary the data generation process in five important aspects to create a comprehensive synthetic experiment under these conditions.
\begin{enumerate}
    \item Number of pre-treatment covariates.
    \item Number of total timesteps.
    \item Number of missing timesteps.
    \item Size of the action space.    
    \item Policy creation method (i.e. how we generate $\pi_i$).
\end{enumerate}

For each of these five aspects, we consider two separate settings. \textbf{We enumerate over all possible combinations for a total of 32 experimental setups.} To align with our real-world dataset size, we set the number of patients $n = 1000$ for all setups. We outline the two options for each aspect below.

\begin{enumerate}
    \item Number of pre-treamtent covariates.
    \begin{enumerate}
        \item 10 pre-treatment covariates ($p = 10$).
        \item 100 pre-treatment covariates ($p = 100$).
    \end{enumerate}
    \item Number of total timesteps.
    \begin{enumerate}
        \item Each patient has two total timesteps ($\tau_i = 2$ for all $i$).
        \item Each patient has between 10 and 15 total timesteps ($T_{min} = 10$, $T_{max} = 15$).
    \end{enumerate}   
    \item Number of missing timesteps.
    \begin{enumerate}
        \item No missing timesteps for any patients ($T_i = \tau_i$ for all $i$).
        \item Patients are missing a variable number of timesteps. If the number of total timesteps is 2(a), then patients are missing between zero and one timesteps ($M_{min} = 0$, $M_{max} = 1$). Otherwise, if the total number of timesteps is 2(b), then patients are missing between two and five timesteps ($M_{min} = 2$, $M_{max} = 5$)
    \end{enumerate}   
    \item Size of the action space.
    \begin{enumerate}
        \item A continuous action space with drug doses allowed in $[0,100]$.
        \item A binary action space with only two drug doses allowed $\{0, 50\}$.
    \end{enumerate}  
    \item Policy creation method (i.e. how we generate $\pi_i$).
    \begin{enumerate}
        \item Random policy. If the action space is continuous, 4(a), then $\pi_i\left(\{E_{i,t'}\}_{t'=1}^{t-1},\{Z_{i,t'}\}_{t'=1}^{t-1} \right) =  \epsilon_{\pi_{i,t}}$ where $\epsilon_{\pi_{i,t}} \sim \text{Uniform}(0,100)$. If the action space is binary, 4(b), then $\pi_i\left(\{E_{i,t'}\}_{t'=1}^{t-1},\{Z_{i,t'}\}_{t'=1}^{t-1} \right) =  50\epsilon_{\pi_{i,t}}$ where $\epsilon_{\pi_{i,t}} \sim \text{Bernoulli}(0.5)$.
        \item An informed policy that is an additive model using ten binary features $F^1,\dots, F^{10}$. For a patient $i$ at timestep $t$, the ten features are calculated as:
        \begin{enumerate}
            \item $F^1_{i, t} = \mathbf{1}[E_{i,t-1} > 10$]
            \item $F^2_{i, t} = \mathbf{1}[E_{i,t-1} > 20]$
            \item $F^3_{i, t} = \mathbf{1}[E_{i,t-1} > 30]$
            \item $F^4_{i, t} = \mathbf{1}[E_{i,t-1} > 40]$
            \item $F^5_{i, t} = \mathbf{1}[E_{i,t-1} > 60]$
            \item $F^6_{i, t} = \mathbf{1}[E_{i,t-1} > 80]$
            \item $F^7_{i, t} = \mathbf{1}[Z_{i,t-1} > 25]$
            \item $F^8_{i, t} = \mathbf{1}[Z_{i,t-1} > 50]$
            \item $F^9_{i, t} = \mathbf{1}[t \geq 3]\mathbf{1}[E_{i,t-1} > 40]\mathbf{1}[\frac{1}{3}\sum_{t'=t-3}^{t-1} E_{i, t'} > 20]$
            \item $F^{10}_{i, t} = \mathbf{1}[t \geq 3]\mathbf{1}[Z_{i,t-1} > 40]\mathbf{1}[\frac{1}{3}\sum_{t'=t-3}^{t-1} Z_{i, t'} > 20]$
        \end{enumerate}
        Then, $\pi_i\left(\{E_{i,t'}\}_{t'=1}^{t-1},\{Z_{i,t'}\}_{t'=1}^{t-1} \right) = \pi_{i}\left(\{F^j_{i,t}\}_{j=1}^{10}\right) = \sum_{j=1}^{10} \zeta_j F^j_{i,t}$, where $\zeta_1,\dots, \zeta_{10}$ are determined by the type of policy assigned to patient $i$. We define three separate policy types: aggressive ($\pi^{a}_{i}$), moderate ($\pi^{m}_{i}$), and conservative ($\pi^{c}_{i}$). Depending on the size of the action space, the coefficients corresponding to each of the policy types are shown in Table~3.

        We then assign a policy to each patient $i$ such that if the patient has a larger average value in $X_{i,1} \text{ and } X_{i,2}$ then they are assigned an aggressive policy with high probability. And similarly, if the patient has a larger average value in $X_{i,3} \text{ and } X_{i,4}$ then they are assigned a conservative policy with high probability. 

        Finally, to emulate a doctor occasionally deviating from the informed policy, at each timestep there is a small chance that the administered dose does not follow the assigned policy $\pi_i$. In particular, if the action space is continuous, 4(a), there is a 5\% chance that $\Z_{i,t} = \xi_{i,t}$ where $\xi_{i,t} \sim \text{Normal}(E_{i,t}, 10)$. And if the action space is binary, 4(b), there is a 5\% chance that $\Z_{i,t} = 50\xi_{i,t}$ where $\xi_{i,t} \sim \text{Bernoulli}(0.5)$.        
    \end{enumerate}      
\end{enumerate}

\begin{table}[]\label{tab: pol-coef-values}
\centering
\begin{tabular}{|
>{\columncolor[HTML]{C0C0C0}}c 
>{\columncolor[HTML]{FFFFFF}}c |
>{\columncolor[HTML]{FFFFFF}}c 
>{\columncolor[HTML]{FFFFFF}}c |
>{\columncolor[HTML]{FFFFFF}}c 
>{\columncolor[HTML]{FFFFFF}}c |
>{\columncolor[HTML]{FFFFFF}}c 
>{\columncolor[HTML]{FFFFFF}}c |}
\hline
\multicolumn{2}{|c|}{\cellcolor[HTML]{C0C0C0}\textbf{Policy Type}}                                                                                    & \multicolumn{2}{c|}{\cellcolor[HTML]{FFFFFF}\textbf{Aggressive}}                           & \multicolumn{2}{c|}{\cellcolor[HTML]{FFFFFF}\textbf{Moderate}}                              & \multicolumn{2}{c|}{\cellcolor[HTML]{FFFFFF}{\color[HTML]{333333} \textbf{Conservative}}}   \\ \hline
\multicolumn{2}{|c|}{\cellcolor[HTML]{C0C0C0}\textbf{Action Space}}                                                                                   & \multicolumn{1}{c|}{\cellcolor[HTML]{FFFFFF}\textbf{Continuous}}         & \textbf{Binary} & \multicolumn{1}{c|}{\cellcolor[HTML]{FFFFFF}\textbf{Continuous}}          & \textbf{Binary} & \multicolumn{1}{c|}{\cellcolor[HTML]{FFFFFF}\textbf{Continuous}}          & \textbf{Binary} \\ \hline
\multicolumn{1}{|c|}{\cellcolor[HTML]{C0C0C0}}                                                                                         & $\zeta_1$    & \multicolumn{1}{c|}{\cellcolor[HTML]{FFFFFF}10+$\epsilon_{\zeta_{a1}}$}  & 0               & \multicolumn{1}{c|}{\cellcolor[HTML]{FFFFFF}$\epsilon_{\zeta_{m1}}$}      & 0               & \multicolumn{1}{c|}{\cellcolor[HTML]{FFFFFF}$\epsilon_{\zeta_{c1}}$}      & 0               \\ \cline{2-8} 
\multicolumn{1}{|c|}{\cellcolor[HTML]{C0C0C0}}                                                                                         & $\zeta_2$    & \multicolumn{1}{c|}{\cellcolor[HTML]{FFFFFF}10+$\epsilon_{\zeta_{a2}}$}  & 50              & \multicolumn{1}{c|}{\cellcolor[HTML]{FFFFFF}$\epsilon_{\zeta_{m2}}$}      & 0               & \multicolumn{1}{c|}{\cellcolor[HTML]{FFFFFF}$\epsilon_{\zeta_{c2}}$}      & 0               \\ \cline{2-8} 
\multicolumn{1}{|c|}{\cellcolor[HTML]{C0C0C0}}                                                                                         & $\zeta_3$    & \multicolumn{1}{c|}{\cellcolor[HTML]{FFFFFF}20+$\epsilon_{\zeta_{a3}}$}  & 0               & \multicolumn{1}{c|}{\cellcolor[HTML]{FFFFFF}10+$\epsilon_{\zeta_{m3}}$}   & 0               & \multicolumn{1}{c|}{\cellcolor[HTML]{FFFFFF}$\epsilon_{\zeta_{c3}}$}      & 0               \\ \cline{2-8} 
\multicolumn{1}{|c|}{\cellcolor[HTML]{C0C0C0}}                                                                                         & $\zeta_4$    & \multicolumn{1}{c|}{\cellcolor[HTML]{FFFFFF}20+$\epsilon_{\zeta_{a4}}$}  & 0               & \multicolumn{1}{c|}{\cellcolor[HTML]{FFFFFF}10+$\epsilon_{\zeta_{m4}}$}   & 0               & \multicolumn{1}{c|}{\cellcolor[HTML]{FFFFFF}$\epsilon_{\zeta_{c4}}$}      & 0               \\ \cline{2-8} 
\multicolumn{1}{|c|}{\cellcolor[HTML]{C0C0C0}}                                                                                         & $\zeta_5$    & \multicolumn{1}{c|}{\cellcolor[HTML]{FFFFFF}20+$\epsilon_{\zeta_{a5}}$}  & 0               & \multicolumn{1}{c|}{\cellcolor[HTML]{FFFFFF}20+$\epsilon_{\zeta_{m5}}$}   & 50              & \multicolumn{1}{c|}{\cellcolor[HTML]{FFFFFF}10+$\epsilon_{\zeta_{c5}}$}   & 50              \\ \cline{2-8} 
\multicolumn{1}{|c|}{\cellcolor[HTML]{C0C0C0}}                                                                                         & $\zeta_6$    & \multicolumn{1}{c|}{\cellcolor[HTML]{FFFFFF}20+$\epsilon_{\zeta_{a6}}$}  & 0               & \multicolumn{1}{c|}{\cellcolor[HTML]{FFFFFF}20+$\epsilon_{\zeta_{m6}}$}   & 0               & \multicolumn{1}{c|}{\cellcolor[HTML]{FFFFFF}20+$\epsilon_{\zeta_{c6}}$}   & 0               \\ \cline{2-8} 
\multicolumn{1}{|c|}{\cellcolor[HTML]{C0C0C0}}                                                                                         & $\zeta_7$    & \multicolumn{1}{c|}{\cellcolor[HTML]{FFFFFF}$\epsilon_{\zeta_{a7}}$}     & 0               & \multicolumn{1}{c|}{\cellcolor[HTML]{FFFFFF}-10+$\epsilon_{\zeta_{m7}}$}  & 0               & \multicolumn{1}{c|}{\cellcolor[HTML]{FFFFFF}-10+$\epsilon_{\zeta_{c7}}$}  & -50             \\ \cline{2-8} 
\multicolumn{1}{|c|}{\cellcolor[HTML]{C0C0C0}}                                                                                         & $\zeta_8$    & \multicolumn{1}{c|}{\cellcolor[HTML]{FFFFFF}$\epsilon_{\zeta_{a8}}$}     & 0               & \multicolumn{1}{c|}{\cellcolor[HTML]{FFFFFF}-20+$\epsilon_{\zeta_{m8}}$}  & 0               & \multicolumn{1}{c|}{\cellcolor[HTML]{FFFFFF}-20+$\epsilon_{\zeta_{c8}}$}  & 0               \\ \cline{2-8} 
\multicolumn{1}{|c|}{\cellcolor[HTML]{C0C0C0}}                                                                                         & $\zeta_9$    & \multicolumn{1}{c|}{\cellcolor[HTML]{FFFFFF}20 +$\epsilon_{\zeta_{a9}}$} & 0               & \multicolumn{1}{c|}{\cellcolor[HTML]{FFFFFF}20 +$\epsilon_{\zeta_{m9}}$}  & 0               & \multicolumn{1}{c|}{\cellcolor[HTML]{FFFFFF}20 +$\epsilon_{\zeta_{c9}}$}  & 0               \\ \cline{2-8} 
\multicolumn{1}{|c|}{\multirow{-10}{*}{\cellcolor[HTML]{C0C0C0}\textbf{\begin{tabular}[c]{@{}c@{}}Coefficient\\ Values\end{tabular}}}} & $\zeta_{10}$ & \multicolumn{1}{c|}{\cellcolor[HTML]{FFFFFF}$\epsilon_{\zeta_{a10}}$}    & 0               & \multicolumn{1}{c|}{\cellcolor[HTML]{FFFFFF}-20+$\epsilon_{\zeta_{m10}}$} & 0               & \multicolumn{1}{c|}{\cellcolor[HTML]{FFFFFF}-20+$\epsilon_{\zeta_{c10}}$} & 0               \\ \hline
\end{tabular}
\caption{Coefficient values for aggressive, moderate, and conservative policies. All $\epsilon_{\zeta_{*}} \overset{iid}{\sim}\text{Normal}(0, 1)$ and are added to emulate the liberty that experts take to slightly deviate from the preset policies.}
\end{table}

Varying these five aspects of the data generation process, we generate a suite of results that provide a comprehensive analysis of the strengths and weaknesses of a variety of optimal policy estimation methods. We outline the methods we compare to, and provided implementation details, in Appendix~\ref{sec: appendix_comp_method}. Results for all experiments are shown in Appendix~\ref{sec: appendix_synth_exp}.

\section{\textsc{Comparison Methods and Implementation Details}}\label{sec: appendix_comp_method}
We compare our matching method to \textit{Finite Timestep Backward Induction Methods}, \textit{Infinite Time Horizon Methods}, and \textit{Deep
Reinforcement Learning Methods}. Many of the methods we compare to are not configured to handle all of the complexities present in our data. For this reason, we make adaptations to each of the methods where necessary. In this section, we outline the methods we implement and any adaptations we make. We omit censored data methods due to their focus on survival analysis time-to-event tasks. We also omit the matching method of \cite{zhou2017causal} as they do not consider multiple timesteps and only discuss discrete treatment options.

\textit{Note One: Many of the methods we compare to can only handle binary or discrete actions spaces. For binary action space methods, we let $Z_{i,t}\in\{0, 50\}$ and we binarize the doses such that $Z_{i,t} = 50\left(\mathbf{1}\left[Z_{i,t} > 25\right]\right)$. For discrete action space methods, we let $Z_{i,t}\in\{0, 25, 50, 75, 100\}$ and we discretize the doses such that $Z_{i,t} = 25\left(\mathbf{1}\left[Z_{i,t} > 12.5\right] + \mathbf{1}\left[Z_{i,t} > 37.5\right] + \mathbf{1}\left[Z_{i,t} > 62.5\right] + \mathbf{1}\left[Z_{i,t} > 87.5\right]\right)$. }

\textit{Note Two: The optimal treatment regime estimation literature normally focuses on maximizing outcomes, not minimizing like we do in our setup. We flip the outcomes in our data for methods that try to maximize in order to account for this.}

\textit{Note Three: A number of the methods we compare to require a reward value corresponding to each patient $i$ at timestep $t$, $\{R_{i,t'}\}_{t'=1}^{T_i}$. To calculate these values, we define three separate reward functions: naive, insightful, and oracle. The naive reward function prioritizes reducing EA burden while avoiding large drug doses, but does not consider the patient's pre-treatment covariates. The insightful reward function considers the interaction between $X_{i,1}$ and EA burdens and $X_{i,3}$ and drug doses, but assumes a linear relationship and does not account for $X_{i,2}$ nor $X_{i,4}$. The oracle reward function is of the same form as our outcome function defined in Equation~\ref{eq: out-func}. We compare to three configurations of each method that requires reward values, where each configuration uses reward values calculated from a different reward function. The exact reward functions are outlined below. Note that all methods aim to maximize the reward function.}

{\footnotesize
\begin{gather}
    \text{Naive: } R_{i,t}\left(\{E_{i,t'}\}_{t'=1}^{t},\{Z_{i,t'}\}_{t'=1}^{t} \right) = \left[E_{i,t-1} - E_{i,t}\right] + \frac{50 - Z_{i,t}}{4}
    \\
    \text{Insightful: } R_{i,t}\left(\{E_{i,t'}\}_{t'=1}^{t},\{Z_{i,t'}\}_{t'=1}^{t} \right) =  -\left[\exp{\left(X_{i,1}\right)} E_{i,t} + \exp{\left(X_{i,3}\right)}Z_{i,t} \right]
    \\
    \text{Oracle: } R_{i,t}\left(\{E_{i,t'}\}_{t'=1}^{t},\{D_{i,t'}\}_{t'=1}^{t} \right) =  -\left[\exp{\left(\sum_{j=1}^{2} \frac{X_{i,j}}{2}\right)} \left(\exp{\left(\frac{E_{i,t}}{50}\right)} - 1 \right) + \exp{\left(\sum_{j=3}^{4} \frac{X_{i,j}}{2}\right)} \left( \exp{\left(\frac{D_{i,t}}{50}\right)} - 1 \right) \right]
\end{gather}
}

\begin{itemize}
    \item \textbf{Finite Timestep Backward Induction Methods:} We compare to a wide array of finite timestep backward induction methods. The methods we compare to are: Q-learning \cite{murphy2005generalization, moodie2012q, clifton2020q}, BOWL \citep{zhao2015new}, and optimal classifier \citep{zhang2012estimating}. We used the R package \texttt{DynTxRegime} \citep{holloway2020dyntxregime} to implement each of these methods. These methods all require a discrete treatment space and the \texttt{DynTxRegime} package only handles the binary case. Given that there is a large literature on Q-learning for discrete action spaces with more than two actions, we also implement our own version of Q-learning for multilevel treatments. For these methods, we followed the Q-learning implementation for observational data as outlined by \cite{moodie2012q}.

    Finite timestep backward induction methods assume full observation of all states and actions for each patient and that the number of timesteps for each patient is the same. To implement these methods when patients have varying numbers of observable timesteps, we truncate the state and action space to only include the timesteps for which all samples have observed data, $\hat{T} = \min_{i\in\{1,\dots,n\}} T_i$. We then carry out each method on this subset of the data to generate estimated optimal treatments for timesteps $t\in\{1,\dots, \hat{T}\}$. From here, we use the model generated at the last observed timestep, $\hat{T}$, to estimate optimal treatments for the remaining $t\in\{\hat{T},\dots,\tau_i\}$ for each patient $i$.  
    
    For the binary Q-learning methods implemented using the \texttt{DynTxRegime} R package we run two versions. One where the contrasts model is a linear model and one where the contrasts model is a decision tree model. For both versions, we use a linear model for the main effects component of the outcome regression. This results in two binary Q-learning varieties.

     For the optimal classifier method we also run two versions. One where the contrasts model is a linear model and one where the contrasts model is a decision tree model. For both versions, we use a linear model for the propensity score model and main effects component of the outcome regression. We use a decision tree classifier for the classification model. This results in two optimal classification varieties.
    
    BOWL requires reward values and thus we run a version for each of the three reward functions. We also run a linear kernel and second degree polynomial kernel version of BOWL for each reward function. All versions use a linear model for the propensity score model. This results in six BOWL varieties.

    For the multilevel Q-learning methods, we incorporate the propensity score at each timestep as a term in our Q-function model \citep{moodie2012q}. All propensity scores are estimated with a linear model. We consider three cases: linear model Q-functions, support vector machine with RBF kernels Q-functions, and random forest Q-functions. This results in three multivel Q-learning varieties.

    \textbf{In total, we generate results from 13 varieties of finite timestep backward induction methods.} 

    \item \textbf{Infinite Time Horizon Methods:} We compare to infinite time horizon Q-learning. We implement this method using \textit{Fitted Q-iteration} as outlined in Algorithm 2 of Section 4 of \cite{clifton2020q}. Similar to multilevel backward induction Q-learning, we use a linear model to estimate propensity scores and include them as a term to the Q-funcion. We consider using three different types of models for the Q-functions: linear models, support vector machines with RBF kernels, and random forests. For each model type, we also consider the case of binarizing the doses into $\{0,50\}$ and discretizing the doses into $\{0,25,50,75,100\}$. Finally, infinite horizon methods need a reward for each action, so we run each configuration under each of the three reward functions.

    \textbf{In total, we generate results from 18 varieties of infinite time horizon methods.}     
    
    \item \textbf{Deep Reinforcement Learning Methods:} We compare to Batch Constrained Q-learning (BCQ) \citep{fujimoto2018off}, Conservative Q-learning (CQL) \citep{kumar2020conservative}, Critic Regularized Regression (CRR) \citep{wang2020critic}, Deep Deterministic Policy Gradients (DDPG) \citep{lillicrap2015continuous}, Soft Actor-Critic (SAC) \citep{haarnoja2018soft}, and Twin Delayed Deep Deterministic Policy Gradients (TD3) \citep{fujimoto2018addressing}. We implement these methods using the \texttt{d3rlpy} Python package \citep{d3rlpy}. All of these methods require reward values, so each was run three separate times -- one for each of the three reward functions. We set the number of steps for each model to 10,000 and kept the remaining parameters at their default values.

    \textbf{In total, we generate results from 18 varieties of deep reinforcement learning methods.} 
\end{itemize}

In addition to the optimal treatment regime estimation methods outlined above, we compare to a handful of other baselines. We refer to these as \textbf{preset policies} and outline each of them below.
\begin{itemize}
    \item \textbf{Expert:} This baseline is meant to emulate an educated doctor strictly following the informed policy with no deviation. Here we assign policies to each patient $i$ as done in the informed policy creation method 5(b). However, we remove all the noise we added to 5(b).In particular, $\epsilon_{\eta_*} = 0$ and there is a 0\% chance that the doctor deviates from the assigned policy at each timestep.

    \item \textbf{Random:} Random dosing at each timestep. If the action space is continuous, 4(a), then $Z_{i,t} = \xi_{i,t}$ where $\xi_{i,t} \sim \text{Uniform}(0,100)$. Otherwise, if the action space is binary, 4(b), then $Z_{i,t} = 50\xi_{i,t}$ where $\xi_{i,t}\sim\text{Bernoulli}(0.5)$.

    \item \textbf{Inaction:} No drug is administered to any patients at any timesteps. $Z_{i,t} = 0$ for all $i$ and $t$.

    \item \textbf{Full Dosing:} If the action space is continuous, 4(a), then a dose of 100 is given at every timestep. $Z_{i,t} = 100$ for all $i$ and $t$. If the action space is binary, 4(b), then a dose of 50 is given at every timestep. $Z_{i,t} = 50$ for all $i$ and $t$.
\end{itemize}

We implement our method as outlined in Section~\ref{sec: method}. Since here we know the true underlying PK/PD parameters, we omit Step 1 from our method to ensure a fair comparison. We first estimate each patient's observed regime with a linear model, using the ten features in 5(b) of Appendix~\ref{sec: dgp} as our policy template. We then learn a distance metric with a linear model and use that distance metric to perform nearest neighbors matching. We create matched groups of size five for each patient, where we match to the five closest patients with good outcomes. Finally, we perform linear interpolation over the patients' policies in each matched group to estimate the optimal policy, $\hat{\pi}_i^*$, for each patient $i$.

\section{\textsc{Synthetic Data Experiments: Additional Results and Implementation Details}}\label{sec: appendix_synth_exp}

In Section~\ref{sec: synth} we present just a small selection of the results from our synthetic data experiment. Here we provide all of our results and further implementation details. We give a comprehensive analysis of key findings in Section~\ref{sec: appendix-add-synth-results}. We provide additional experimental implementation details in Section~\ref{sec: appendix-add-synth-details}. Given the number of approaches (54) and data generation process setups (32) we ran tests for, we include our full results in our publically available \href{https://github.com/almost-matching-exactly/opt_tx_regime_matching}{GitHub repository} (https://github.com/almost-matching-exactly/opt\_tx\_regime\_matching). We outline each file and its contents in Section~\ref{sec: appendix-full-synth-results-files}.

\subsection{Additional Results for Synthetic Data Experiments}\label{sec: appendix-add-synth-results}

\paragraph{Summary of our Analysis.} We first compare our method with the 39 approaches that do not use the oracle reward function and are not a preset policy. As noted in Section~\ref{sec: synth}, on the 8 setups with 10-15 timesteps where the observed data is generated from educated policies, our method is consistently the top performer. Looking at the 8 setups where we have 10-15 timesteps and 2-5 missing timesteps, our method outperforms all other approaches in the majority of setups (5 of 8) and is always among the top four performing approaches -- never more than 4.5 percentage points worse than the best approach. In the 16 setups with 10-15 timesteps we are the best performing method 9 of 16 times and among the top 4 approaches 16 of 16 times - never more than 7 percentage points worse than the top approach. When compared on all 32 simulations setups, where some are specifically designed for finite-timestep backward induction methods to perform well, our method outperforms all of the comparison approaches in 17 of the 32 setups and is among the top 4 approaches 29 times -- never more than 10.1 percentage points worse than the top approach.

When we also consider the oracle reward functions, our method is never more than 12 percentage points worse than the top performing approach on the 8 setups with 10-15 timesteps and 2-5 missing timesteps and never more than 15 percentage points worse than the top performing approach across all 32 setups.

All of these upper limits on the number of percentage points between our method and the top performing approach are the lowest such values for any method. Ultimately, our simulation results show that our method is frequently the best approach and that its performance is consistent across a variety of scenarios.

In the remainder of this section, we perform an in-depth analysis comparing our method to each of the categories of existing DTR and RL methods that we implemented. We focus on finite-timestep backward induction methods Q-learning and optimal classifier in Appendix~\ref{sec: appendix_qlearn_optclass}, infinite horizon methods in Appendix~\ref{sec: appendix_infhorizon}, Deep RL in Appendix~\ref{sec: appendix_deeprl}, and BOWL in Appendix~\ref{sec: appendix_bowl}. In each subsection, we comment on the strengths and weaknesses of the methods, ultimately highlighting how our approach is superior for estimating optimal treatment regimes in complex high-stakes settings.

\subsubsection{Analyzing \textbf{Q-learning} and \textbf{Optimal Classifier} Performance}\label{sec: appendix_qlearn_optclass}
\begin{figure}[h]
    \centering
    \includegraphics[width=0.9\linewidth]{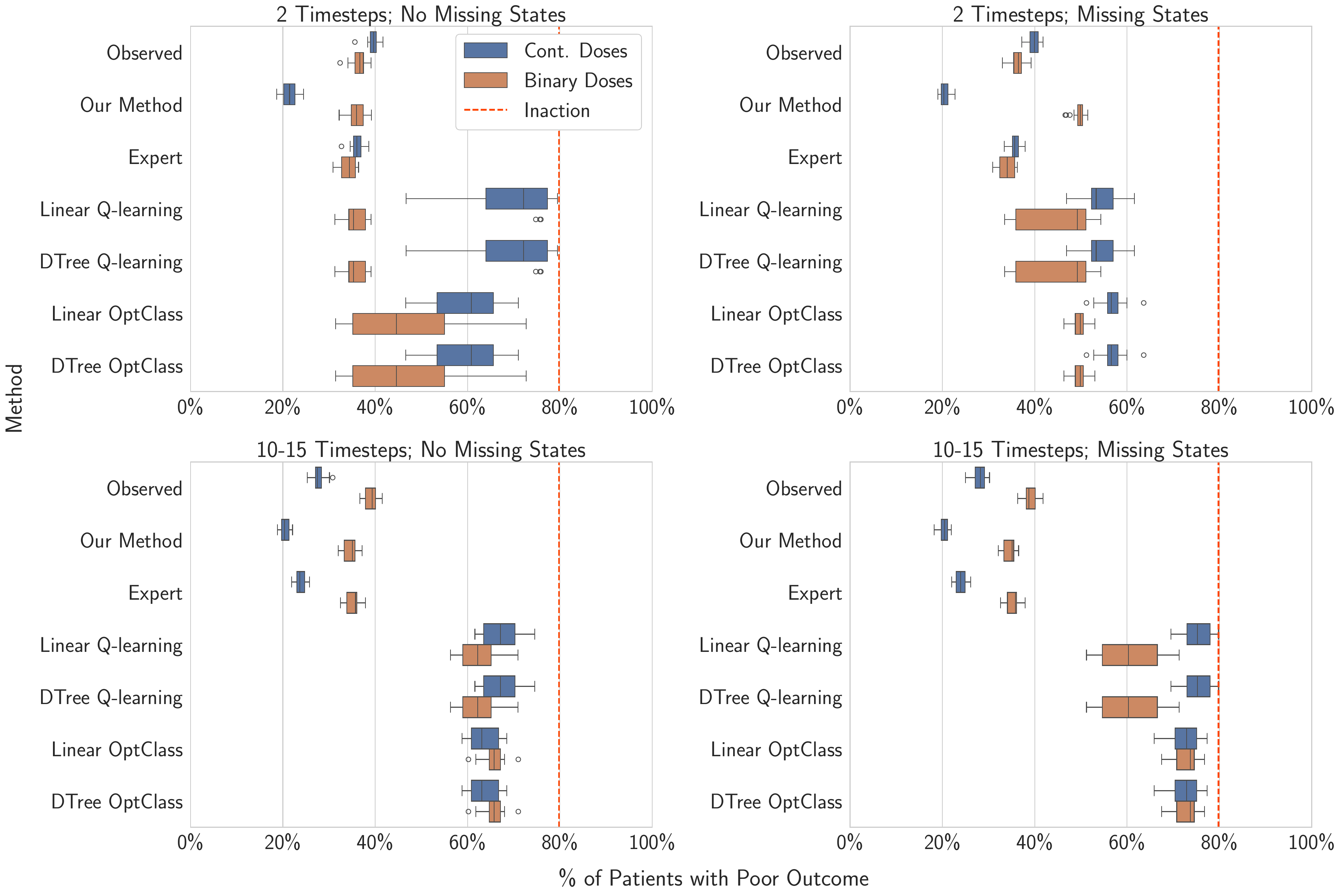}
    \caption{Percent of patients with poor outcomes under different proposed policies (\textit{lower is better}). Boxplots show the distribution of the average outcomes over 20 iterations. \textit{Observed} shows average observed outcomes. \textit{Expert} shows outcomes under the expert policies. \textit{Linear} and \textit{DTree Q-learning} are finite-timestep backward induction Q-learning using either linear models or decision trees. \textit{Linear} and \textit{DTree OptClass} are optimal classifier using either linear models or decision trees. See Appendix~\ref{sec: appendix_comp_method} for further details of each method. Note that not all backward induction methods converged for all 20 iterations of each setup. See \texttt{all\_sims\_nan.csv} and Appendix~\ref{sec: appendix-full-synth-results-files} for details.}
    \label{fig: binary-comp1}
\end{figure}

 \textbf{Q-learning} and \textbf{Optimal Classifier} methods implemented using the \texttt{DynTxRegime} R package struggle in complex settings for what we presume is a variety of reasons. Figure~\ref{fig: binary-comp1} details the performance of our method, Q-learning, and optimal classifier with varying action spaces (binary vs. continuous), number of timesteps (2 vs. 10-15), and missing states (missing vs. no missing). These plots highlight how our method drastically outperforms Q-learning and optimal classifier in continuous action spaces. It makes sense that Q-learning and optimal classifier struggle with continuous actions spaces, as they are forced to binarize continuous actions spaces, thereby losing important information. Note that the best results across all the plots in Figure~\ref{fig: binary-comp1} are achieved by our method when we allow the doses to be continuous, suggesting that binarizing the treatment is not a good strategy to optimize outcomes for patients. 
 
 We also note that our method is far superior in settings with longer time horizons. This aligns with the fact that previous work on finite-timestep backward induction methods has largely focused on the two timestep setting, paying less attention to longer time horizons \citep{clifton2020q}. As outlined in Appendix~\ref{sec: appendix_comp_method}, when implementing these methods we truncate all of the states to only include timesteps for which all individuals have an observed state and action. This removes a large amount of information from the data and most likely impacts the performance of these methods. 

 We further note that the finite-timestep backward induction methods perform better, on average, when there are no missing timesteps. Whereas, our method is quite robust to the missingness of states.

\begin{figure}[h]
    \centering
    \includegraphics[width=0.45\linewidth]{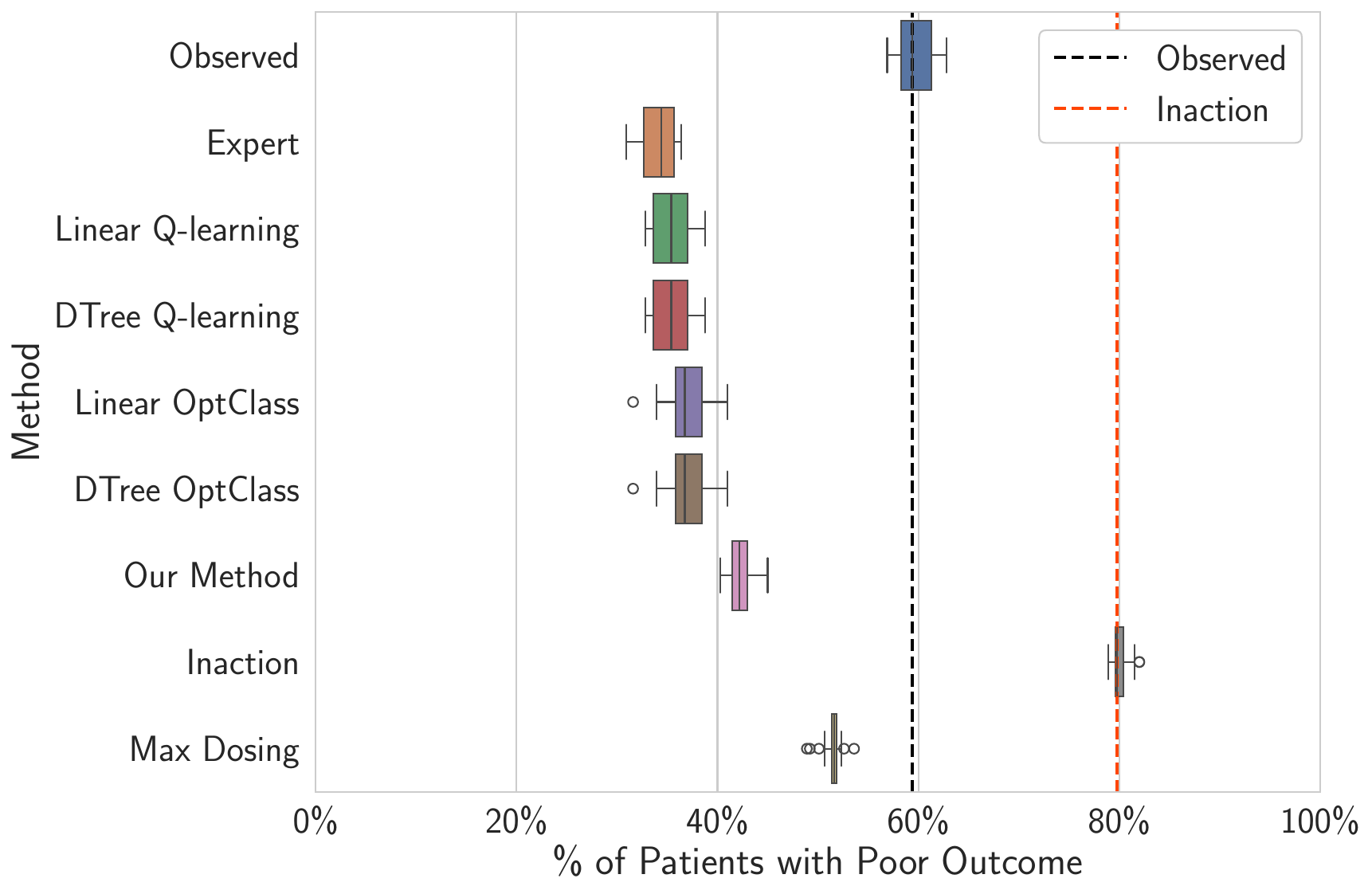}
    \caption{Percent of patients with poor outcomes under different proposed policies (\textit{lower is better}) in a setting more conducive to finite-timestep backward induction methods. Here we set the (i) number of covariates to 10, (ii) number of timesteps to 2, (iii) have no missing states, (iv) only allow binary doses, and (v) generate the observed data from a random policy. Boxplots show the distribution of the average outcomes over 20 iterations. \textit{Observed} shows average observed outcomes. \textit{Expert} shows outcomes under expert policies. \textit{Inaction} and \textit{Max Dosing} administer no drugs and the max amount of drugs to each patient at each timestep, respectively. \textit{Linear} and \textit{DTree Q-learning} are finite-timestep backward induction Q-learning using either linear models or decision trees. \textit{Linear} and \textit{DTree OptClass} are optimal classifier using either linear models or decision trees. See Appendix~\ref{sec: appendix_comp_method} for further details of each method.}
    \label{fig: binary-sanity}
\end{figure}

As a sanity check, we show the performance of Q-learning and optimal classifier under the conditions that it was primarily designed for in Figure~\ref{fig: binary-sanity}. These results show how effective Q-learning and optimal classifer can be in a more conducive setting, with all varieties outperforming our method. However, this performance does not translate to our challenging high-stakes setting, ultimately making these methods ill-suited for our application.

One obvious way to try to improve finite-timestep backward induction Q-learning is to decrease the amount of information loss by discretizing the continuous treatments into more bins. Since the \texttt{DynTxRegime} R package does not support multilevel treatments we implemented our own version of Q-learning to handle this. We outline our implementation in Appendix~\ref{sec: appendix_comp_method}. 

Figure~\ref{fig: binary-vs-multi} shows a comparison between binary Q-learning with two treatment options and discrete Q-learning with five treatment options. All plots in Figure~\ref{fig: binary-vs-multi} are in settings with 10 pre-treatment covariates, a continuous action space, and observed data generated using an informed policy. While we do see a gain in performance, this gain is less substantial when there are more timesteps and missing states. Ultimately, the multi-level treatment form of Q-learning still fails to match the performance of our method. This suggests that while Q-learning can improve by increasing the number of discrete dose options, it still struggles with long time horizons and missing states. Furthermore, at some point the small sample size limits the gain in performance Q-learning can achieve by creating more treatment bins.

\begin{figure}[h]
    \centering
    \includegraphics[width=0.9\linewidth]{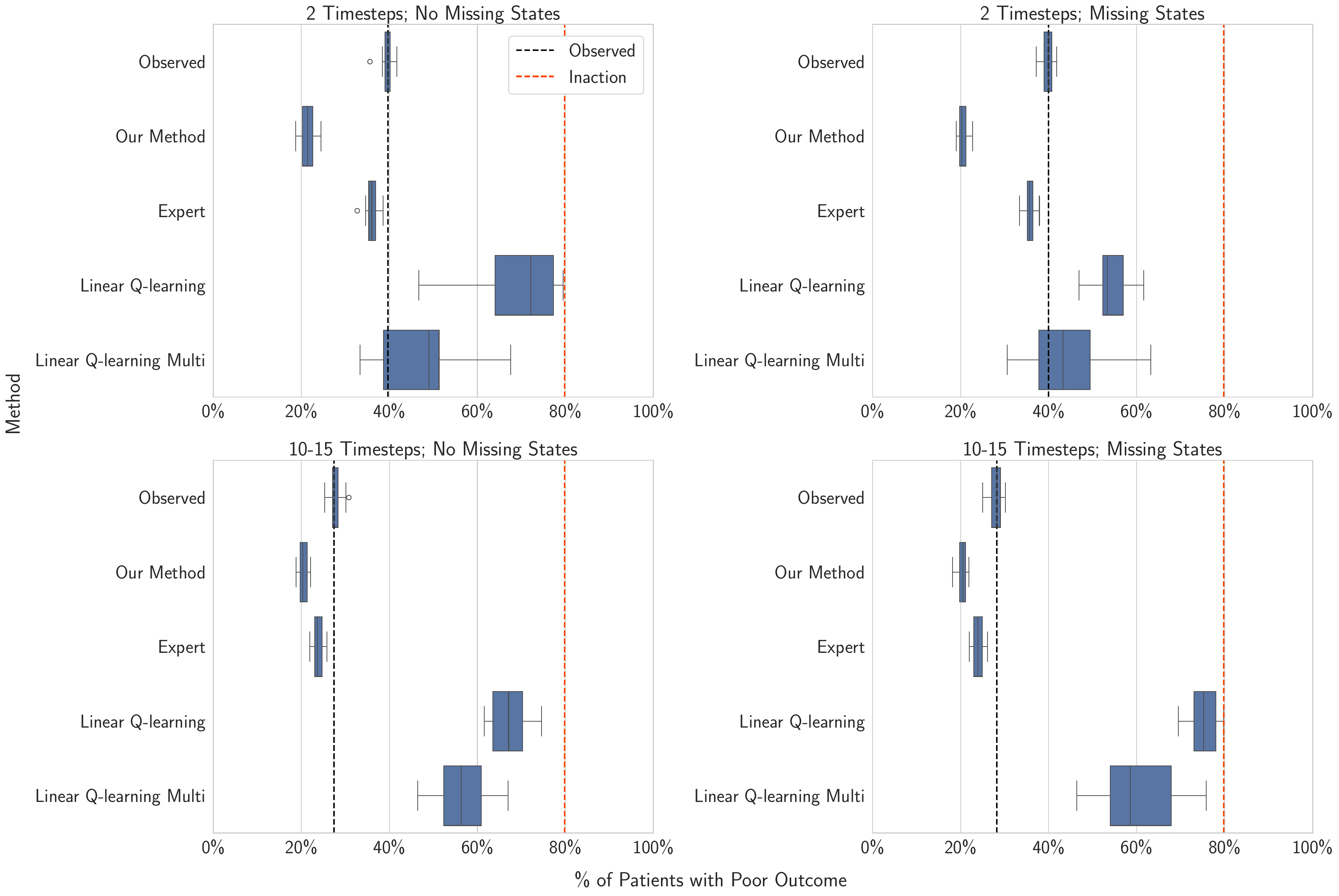}
    \caption{Percent of patients with poor outcomes under different proposed policies (\textit{lower is better}). In all plots, the setup has (i) 10 pre-treatment covariates, (iv) a continuous action space, and (v) generates the observed data using an informed policy. Boxplots show the distribution of the average outcomes over 20 iterations. \textit{Observed} shows average observed outcomes. \textit{Expert} shows outcomes under expert policies. \textit{Linear Q-learning} and \textit{Linear Q-learning Multi}  are finite-timestep backward induction Q-learning using linear models. \textit{Linear Q-learning} binarizes the continuous treatment values into two values whereas \textit{Linear Q-learning Multi} discretizes the treatments into five bins. See Appendix~\ref{sec: appendix_comp_method} for further details of each method.}
    \label{fig: binary-vs-multi}
\end{figure}

\subsubsection{Analyzing Infinite Horizon Performance}\label{sec: appendix_infhorizon}
\begin{figure}[h]
    \centering
    \includegraphics[width=0.9\linewidth]{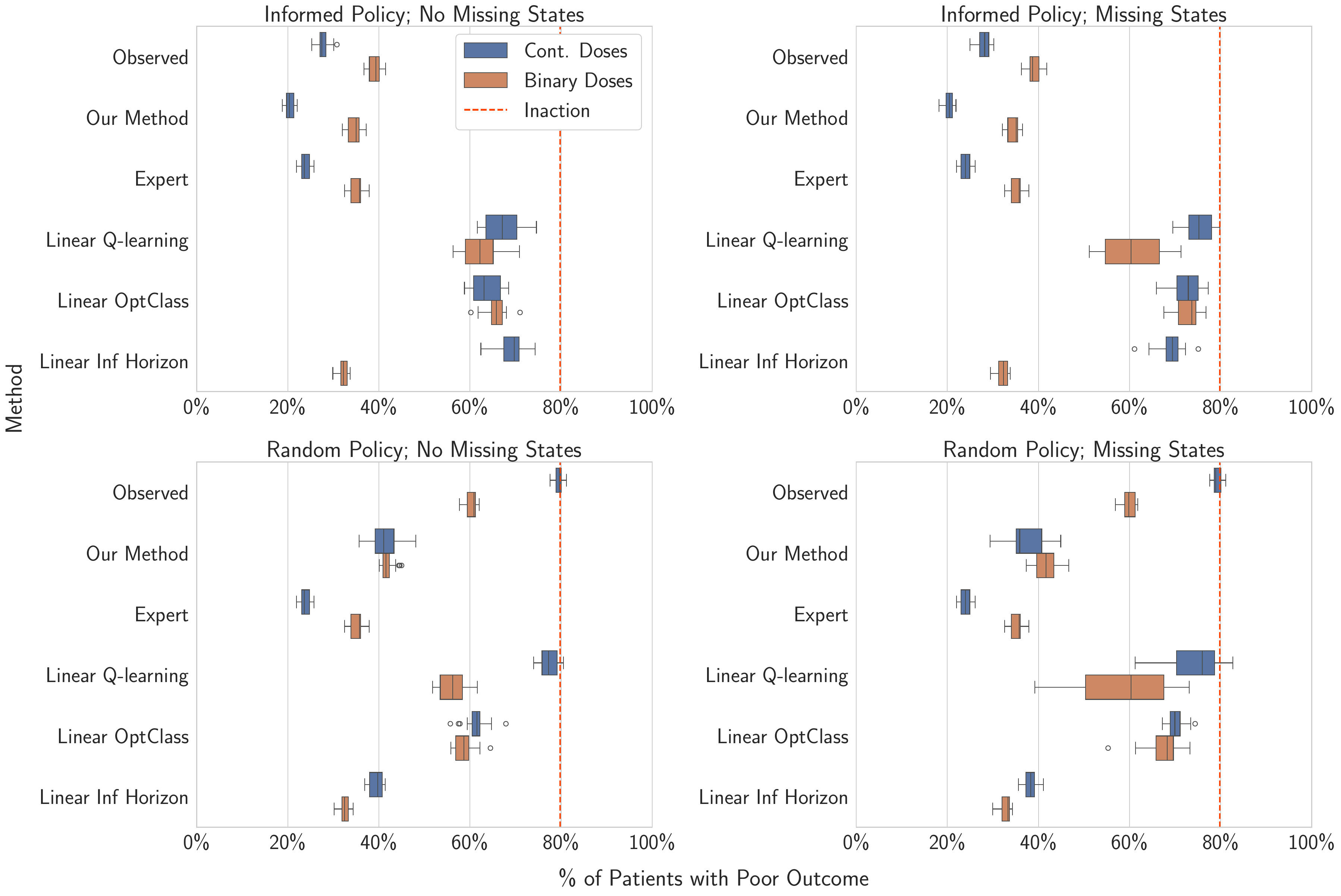}
    \caption{Percent of patients with poor outcomes under different proposed policies (\textit{lower is better}). In all plots, the setup has (i) 10 pre-treatment covariates and (ii) 10-15 total timesteps. Boxplots show the distribution of the average outcomes over 20 iterations. \textit{Observed} shows average observed outcomes. \textit{Expert} shows outcomes under expert policies. \textit{Linear Q-learning} is finite-timestep backward induction Q-learning using linear models. \textit{Linear OptClass} is optimal classifier using linear models. \textit{Linear Inf} is the infinite horizon method fitted Q-iteration using linear models. \textit{Linear Inf} uses the oracle reward function. See Appendix~\ref{sec: appendix_comp_method} for further details of each method and the reward functions.}
    \label{fig: inf_vs_finite}
\end{figure}

\textbf{Infinite Horizon} methods can overcome the issue finite-timestep backward induction methods face with longer time horizons and missing states/actions. Figure~\ref{fig: inf_vs_finite} shows a comparison of our method, the infinite horizon method fitted Q-iteration (see \cite{clifton2020q}), and the finite-timestep backward induction methods Q-learning and optimal classifier. The subplots in this figure show how each method performs with different numbers of missing states and different size action spaces.

Figure~\ref{fig: inf_vs_finite} highlights how infinite horizon methods can handle long time horizons and missing states much better than finite-timestep backward induction Q-learning and optimal classifier. Fitted Q-iteration can outperform our method when the action space is binary and does especially well when the observed data is generated from a random policy. 

However, we still see that fitted Q-iteration struggles with a continuous action space. This is particularly true when the observed data is generated from an informed policy (first row plots of Figure~\ref{fig: inf_vs_finite}). Conversely, our method can handle these added complexities, producing much better results in the setups most resembling a complex real-world setting.

\begin{figure}[h]
    \centering
    \includegraphics[width=0.9\linewidth]{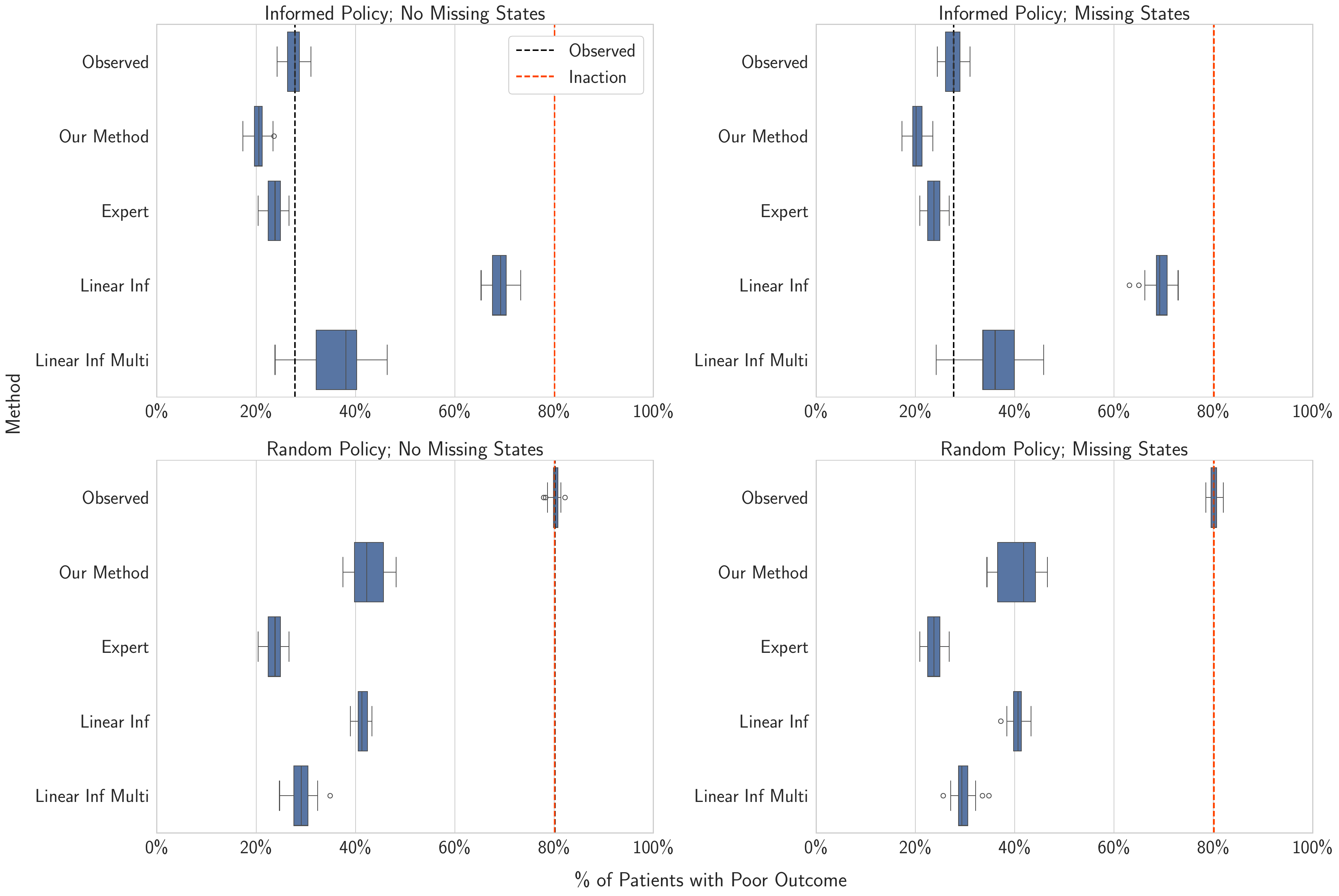}
    \caption{Percent of patients with poor outcomes under different proposed policies (\textit{lower is better}). In all plots, the setup has (i) 100 pre-treatment covariates, (ii) 10-15 total timesteps, and (iv) a continuous action space. Boxplots show the distribution of the average outcomes over 20 iterations. \textit{Observed} shows average observed outcomes. \textit{Expert} shows outcomes under expert policies. \textit{Linear Inf} is Fitted Q-iteration where the treatments are binarized  and \textit{Linear Inf Multi} is Fitted Q-iteration where the treatments are discretized into five bins. Both methods use linear models and the oracle reward function (see Appendix~\ref{sec: appendix_comp_method} for details on reward functions).}
    \label{fig: inf_binary_vs_multi}
\end{figure}

Similar to Figure~\ref{fig: binary-vs-multi} for backward induction Q-learning, Figure~\ref{fig: inf_binary_vs_multi} shows how infinite horizon methods can alleviate the problem of a continuous action space by using a multi-level treatment version of fitted Q-iteration instead of a binary version. The bottom row of Figure~\ref{fig: inf_binary_vs_multi} shows the strong performance of infinite horizon methods when using observational data generated from a random policy. Fitted Q-iteration outperforms our method in these setups. However, when the training data is generated from an informed policy (top row of Figure~\ref{fig: inf_binary_vs_multi}), as observational data typically is, our method has superior performance. This could be due to the fact that infinite horizon methods have to deal with the notion of exploration vs. exploitation \citep{clifton2020q}, leading to worse performance when the data is collected following a relatively stagnant and educated policy. Observed data collected under such policies essentially has less "exploration" built into it. This struggle could also be due to the fact that infinite horizon methods often work under the assumption that there is a non-zero probability of each action at each timestep \citep{ertefaie2018constructing}. However, under the informed policy there are certain states for which certain actions are near-impossible.

\begin{figure}[h]
    \centering
    \includegraphics[width=0.9\linewidth]{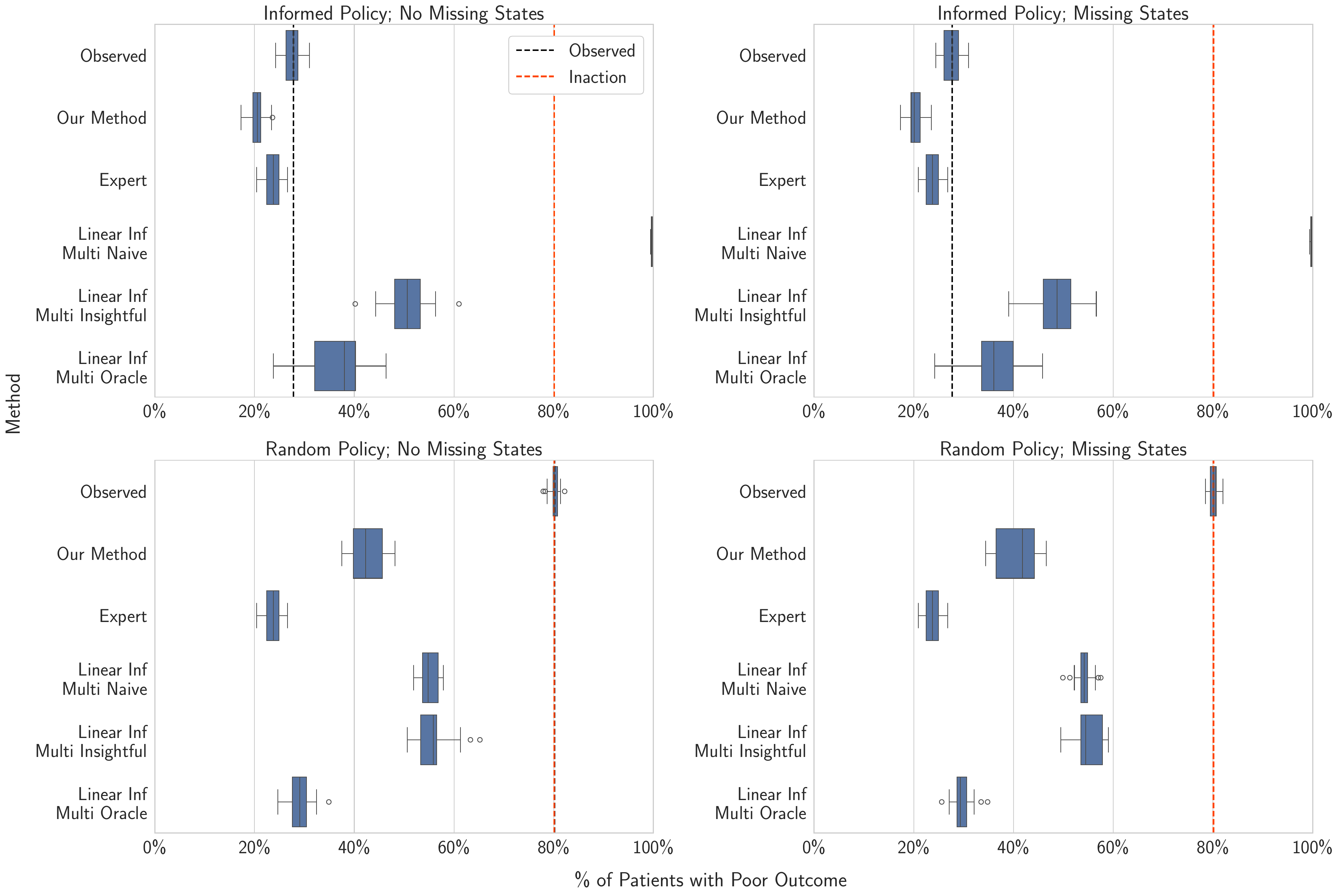}
    \caption{Percent of patients with poor outcomes under different proposed policies (\textit{lower is better}). In all plots, the setup has (i) 100 pre-treatment covariates, (ii) 10-15 total timesteps, and (iv) a continuous action space. Boxplots show the distribution of the average outcomes over 20 iterations. \textit{Observed} shows average observed outcomes. \textit{Expert} shows outcomes under expert policies. All \textit{Linear Inf Multi} methods are fitted Q-iteration approaches that discretize the treatment into five bins and use linear models. \textit{Naive}, \textit{Insightful}, or \textit{Oracle} at the end of each \textit{Linear Inf Multi} method specifies which reward function is uses to calculate the reward values. See Appendix~\ref{sec: appendix_comp_method} for details on reward functions.}
    \label{fig: inf_rewards}
\end{figure}

Infinite horizon methods are a promising technique, but face a key challenge in our data setup as they require a reward value to be assigned to each action. In our setup, we only observe an outcome at the end of a patient's timesteps. Therefore, we are forced to define a reward function ourselves. We outline the three different reward functions we consider in Appendix~\ref{sec: appendix_comp_method}. Figure ~\ref{fig: inf_binary_vs_multi} showed results using the oracle reward function. Figure~\ref{fig: inf_rewards} depicts the stark differences in performance we see using infinite horizon methods with different reward functions. We observe that the performance of infinite horizon methods suffers as the reward function gets farther away from the truth. Researchers typically do not know the oracle, or true, reward function and while we can compare the different reward functions since we know the underlying simulation setup, this is not the case with real-world observational data. Thus, the researcher has to carefully consider the reward function when using infinite horizon methods. This ultimately limits the usefulness of infinite horizon methods in high-stakes applications where reward values are not available for each action that is observed.

\subsubsection{Analyzing Deep RL Performance}\label{sec: appendix_deeprl}
\begin{figure}[h]
    \centering
    \includegraphics[width=0.45\linewidth]{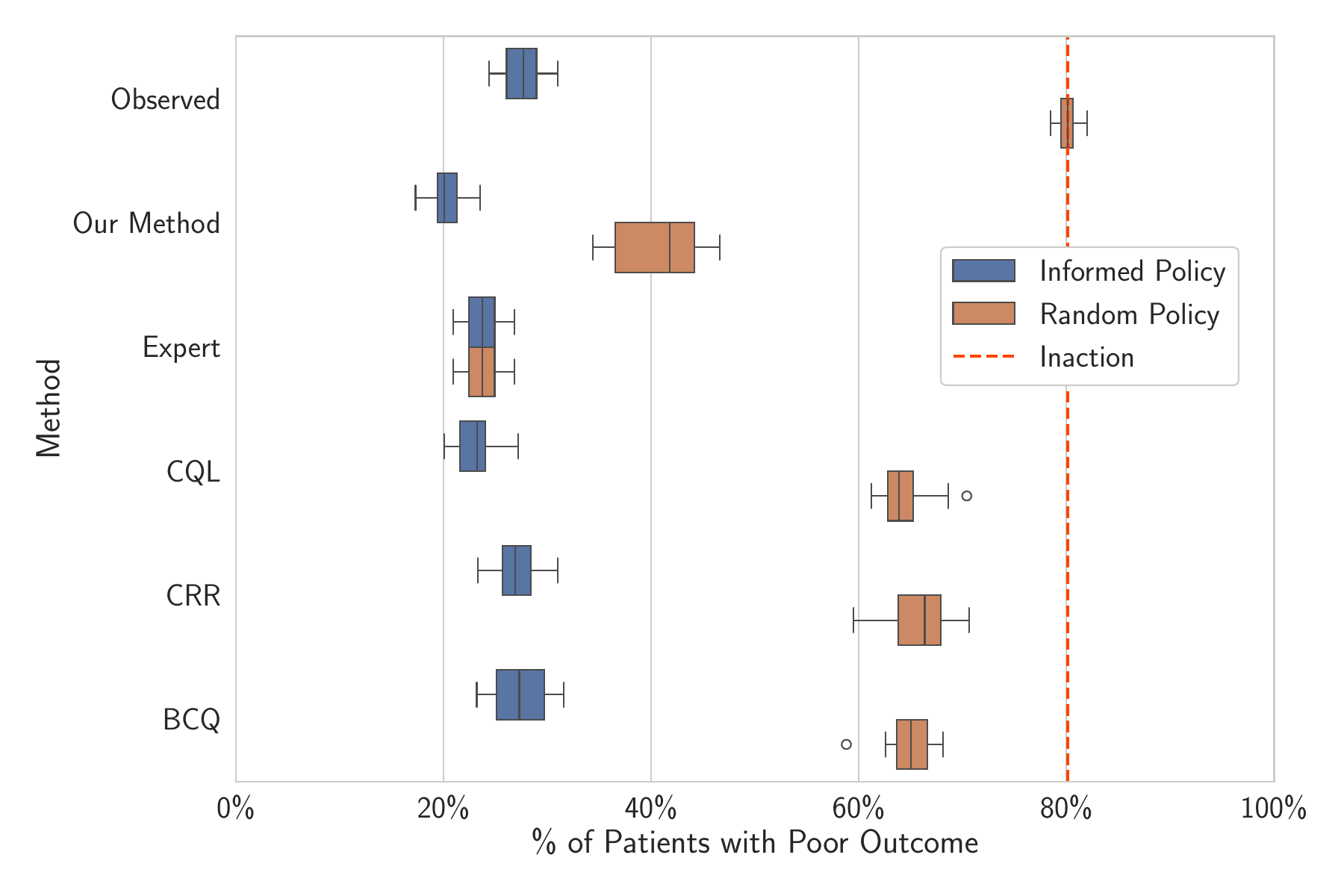}
    \caption{Percent of patients with poor outcomes under different proposed policies (\textit{lower is better}). In all plots, the setup has (i) 100 pre-treatment covariates, (ii) 10-15 total timesteps, (iii) 2-5 missing states, and (iv) a continuous action space. Boxplots show the distribution of the average outcomes over 20 iterations. \textit{Observed} shows average observed outcomes. \textit{Expert} shows outcomes under expert policies. \textit{CQL}, \textit{CRR}, and \textit{BCQ} are all Deep RL methods using the insightful reward function. See Appendix~\ref{sec: appendix_comp_method} for details on reward functions.}
    \label{fig: deep_policy}
\end{figure}

The performance capabilities of \textbf{Deep Reinforcement Learning} is already depicted in Section~\ref{sec: synth}'s Figure~\ref{fig: synthetic-sim1}. While DDPG, SAC, and TD3 struggle with the smaller sample sizes and/or the lack of randomness in informed policies, the more modern architectures like BCQ, CQL, and CRR perform well, although slightly worse than our method, on a simulated dataset that resembles our real-world data. However, we note that these Deep RL methods struggle when a random policy is used to generate the observed data. Figure~\ref{fig: deep_policy} shows how BCQ, CQL, and CRR perform worse when the training data is generated from a random policy. While our method's performance also suffers in this setting, the dip in performance is less severe than Deep RL methods. We hypothesize that Deep RL struggles when using data generated from a random policy because they all use an evaluation set to guide the learning process \citep{d3rlpy}. Thus, with only a small amount of data generated via a random policy, it is difficult to evaluate the model's performance. Deep RL methods would likely improve if we had significantly more randomly generated data or had the ability to do online learning \citep{luo2023finetuning}.

\begin{figure}[h]
    \centering
    \includegraphics[width=0.45\linewidth]{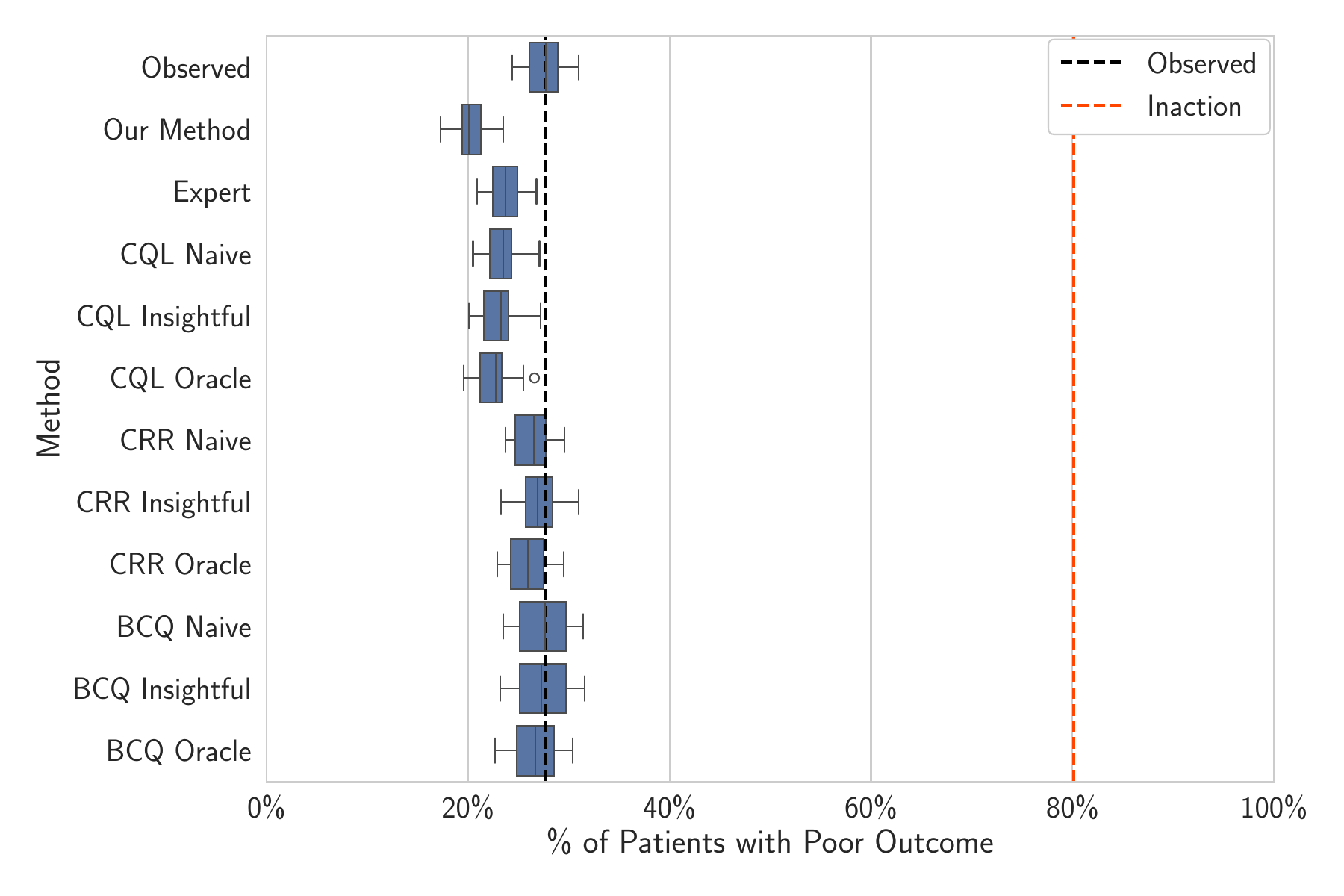}
    \caption{Percent of patients with poor outcomes under different proposed policies (\textit{lower is better}). In all plots, the setup has (i) 100 pre-treatment covariates, (ii) 10-15 total timesteps, (iii) 2-5 missing states, (iv) a continuous action space, and (v) generates data from an informed policy. Boxplots show the distribution of the average outcomes over 20 iterations. \textit{Observed} shows average observed outcomes. \textit{Expert} shows outcomes under expert policies. \textit{CQL}, \textit{CRR}, and \textit{BCQ} are all Deep RL methods where \textit{Naive}, \textit{Insightful}, or \textit{Oracle} at the end of each method specifies which reward function is uses to calculate the reward values. See Appendix~\ref{sec: appendix_comp_method} for details on reward functions.}
    \label{fig: deep_reward}
\end{figure}

Deep RL approaches, like infinite horizon methods, also require a reward to be specified for each action. Figure~\ref{fig: deep_reward} shows that the performance of the best Deep RL methods when using each of the three different reward functions outlined in Appendix~\ref{sec: appendix_comp_method}. We observe that the performance is stable across these three reward functions when training on data generated from informed policies. Although, we note that all three reward functions are at least slightly related to the outcome, and thus performance could suffer if the reward function was badly misspecified.

Ultimately, deep reinforcement learning methods show promise for optimal treatment regime estimation from observational data generated by domain experts. The main drawbacks of Deep RL in our setting is its fundamental lack of interpretability. The inability to explain the estimates generated by Deep RL makes it ill-suited for high-stakes applications in the medical field.

As an aside, we also note that Deep RL methods require substantially more compute power to train than our method and any of the other methods we compare to. We train these models using significantly more compute power and GPUs Even with the enhanced computing power, these methods have substantially longer runtimes. See Appendix~\ref{sec: appendix-add-synth-details} for further details. 

\subsubsection{Analyzing BOWL Performance}\label{sec: appendix_bowl}
\begin{figure}[h]
    \centering
    \includegraphics[width=0.45\linewidth]{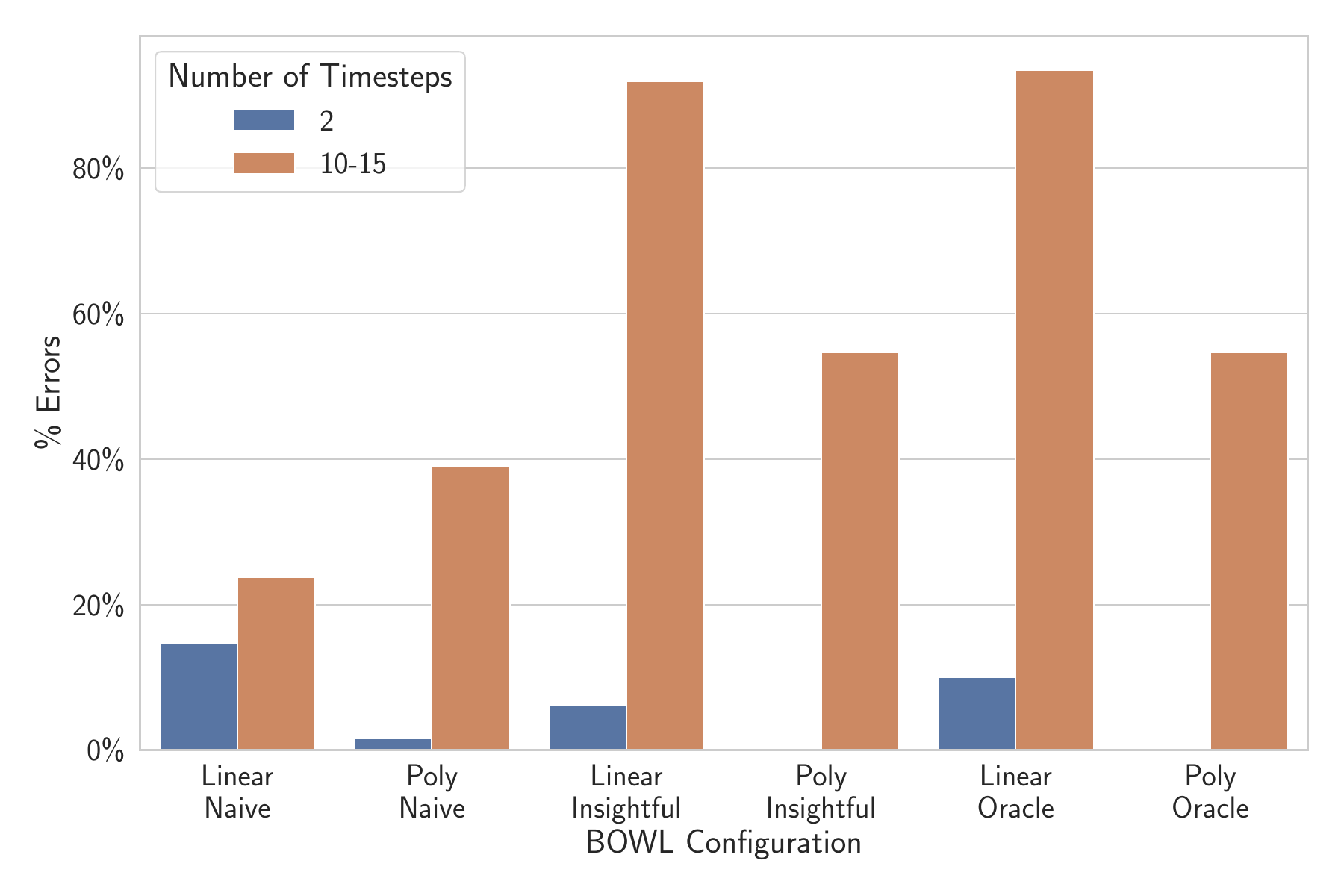}
    \caption{Percentage of total simulation iterations for which different BOWL variations produced either a runtime or convergence error. The different colored bars show the percentage of runs that were errors when the data had either 2 total timesteps or 10-15 total timesteps (see Appendix~\ref{sec: appendix-dgp-setups}). \textit{Linear} or \textit{Poly} refer to the kernel type BOWL uses. \textit{Naive}, \textit{Insightful}, or \textit{Oracle} at the end of each method specifies which reward function is used to calculate the reward values. See Appendix~\ref{sec: appendix_comp_method} for details on BOWL implementation and reward functions. See Appendix~\ref{sec: appendix-add-synth-details} for further details on BOWL and \texttt{DynTxRegime} errors. See \texttt{all\_sim\_nan.csv} and Appendix~\ref{sec: appendix-full-synth-results-files} for full results on which simulation setups BOWL failed to run for.}
    \label{fig: bowl-nan}
\end{figure}

The final method we compare to is Backward Outcome Weighted Learning, \textbf{BOWL}. We find that the BOWL method implemented in \texttt{DynTxRegime} struggles to consistently converge, especially when the training data has more timesteps. We show the frequency in which BOWL fails to run for different configurations and reward functions in Figure~\ref{fig: bowl-nan}. We further discuss the most likely reasons for these runtime issues, and the steps we took to avoid them, in Appendix~\ref{sec: appendix-add-synth-details}. The instability of BOWL for the vast majority of our data configuration setups makes it difficult to discern what aspects of the data are causing it the most problems. We ultimately conclude that BOWL, as implemented in the \texttt{DynTxRegime} R package, is ill equipped to handle the challenges present in our simulated data.

\subsection{Additional Implementation Details for Synthetic Data Experiments}\label{sec: appendix-add-synth-details}
Code to reproduce the results in this paper is available at \href{https://github.com/almost-matching-exactly/opt_tx_regime_matching}{https://github.com/almost-matching-exactly/opt\_tx\_regime\_matching}.

We run each of the methods outlined in Section~\ref{sec: appendix_comp_method} for a total of 20 iterations for each data generation setup. Tests are run on a Slurm cluster with VMware, where each VM is an Intel(R) Xeon(R) Gold CPU (either 5317 @ 3.00GHz, 5320 @ 2.20GHz, 6142 @ 2.60GHz, 6152 @ 2.10GHz, 6226 @ 2.70GHz, or 6252 @ 2.10GHz). Deep RL methods are run on machines with RTX2080 GPUs. Slurm jobs are allocated a single core with 2 GB of RAM for non-Deep RL methods and 16 GB of RAM for Deep RL methods. We always set the random seed to match the iteration number of each setup.

We split the dataset into 5 folds to perform estimation using our method and Deep RL methods. For our method, we use 1 fold to learn the distance metric and perform estimation on the remaining 4 folds. We then average across the 4 outcomes for each sample. For Deep RL methods, we use 4 folds for training and perform estimation on the remaining fold - doing this 5 times to get estimates for each sample.

There are some data generation processes for which we did not generate results for each method for all 20 iterations. You can find details on which methods failed to run for which setups in the \texttt{all\_sims\_nan.csv} file described in Appendix~\ref{sec: appendix-full-synth-results-files}. We outline which methods we are missing results for and provide potential explanations below.
\begin{itemize}
    \item Q-linear, Optimal Classifier, and BOWL implemented using the \textit{DynTxRegime} R package: Each of these methods is missing results for some of the setups because they failed to converge or produced a runtime error. We attempted to alleviate these issues by running both Q-learning and optimal classifier with decision trees and linear models and running BOWL with a linear kernel and a second degree polynomial kernel. We performed five-fold cross-validation to choose the lambda for BOWL. However, the package errored out if any of the folds failed to converge. We added exception handling to account for this, where we attempted to fit BOWL with preset lambda values of 2 and then 0.5 if it produced an error while performing cross-validation.

    After investigation, we hypothesized that Q-learning and optimal classification failed to converge for extremal propensity scores in observational data. This is supported by the fact that their errors only occurred when the policy was semi-random. For this policy choice, there are timesteps where a patient's next dose is mostly predetermined by their current state - thus leading to very small or large propensity scores. 

    We believe that BOWL struggles with a similar issue, given that it also employs the use of a propensity score. However, BOWL failed to converge for a number of the setups that used a random policy. We acknowledge that the software package made it difficult to discern if the errors were being produced due to an issue with how we were implementing it in the \texttt{DynTxRegime} R package or with the BOWL method itself. Thus, we are less sure of the exact reasons why BOWL struggled for so many of our setups.

    \item Multilevel Q-learning and Infinite Horizon methods: We only ran multilevel treatment method when the treatment was continuous, as running these methods when the treatment was binary was equivalent to the binary version of the method.

    \item Deep RL methods implemented using the \texttt{d3rlpy} Python package: The Deep RL methods we compare to only accept continuous actions spaces. Therefore, we do not have results for any of the setups where the action space was discrete. Also, for two of the setups with continuous action spaces, 2 total timesteps, and 0-1 missing timesteps the number of realized doses was such that the methods interpreted the action space as discrete in some of the iterations, causing it to error out.
\end{itemize}

\subsection{Synthetic Data Experiments Results Files}\label{sec: appendix-full-synth-results-files}

We include files with results for all 54 approaches and 32 simulation setups in our public GitHub repository (\href{https://github.com/almost-matching-exactly/opt_tx_regime_matching/tree/main/simulations_data}{https://github.com/almost-matching-exactly/opt\_tx\_regime\_matching/tree/main/simulations\_data}). The files use seven columns to indicate the settings of the data generation process for that run.
\begin{itemize}
    \item \texttt{Sim}: Indicates the assigned simulation number. All rows with the same sim number are run under the same data generation configuration, except for the random seed.
    \item \texttt{Iter}: Indicates the iteration number of the corresponding \texttt{Sim}. The \texttt{Iter} value is also used as the random seed for that run.
    \item \texttt{Covs}: The number of pre-treatment covariates.
    \item \texttt{T Setting}: The number of total timesteps setting, where \texttt{a} corresponds to setup 2(a) and \texttt{b} corresponds to setup 2(b) (see Appendix~\ref{sec: appendix-dgp-setups}).
    \item \texttt{T Drop Setting}: The number of unobserved timesteps setting, where \texttt{a} corresponds to setup 3(a) and \texttt{b} corresponds to setup 3(b) (see Appendix~\ref{sec: appendix-dgp-setups}).
    \item \texttt{Binary Dose}: Whether the treatment space is binary or not (if \texttt{FALSE} then treatment space is continuous).
    \item \texttt{Policy}: The policy used to generate the observed data. If \texttt{random}, than a random policy was used to generate the data. Else if \texttt{informed}, than an informed policy was used (see Appendix~\ref{sec: appendix-dgp-setups}).
\end{itemize}

For all methods that use a reward function, we refer to to the Naive reward function as R1, the insightful reward function as R2, and the oracle reward function as R3. We outline the contents of each file below.
\begin{itemize}
    \item
    \texttt{all\_sims\_binary\_outcomes.csv}: This file contains the average binary outcome value, $\frac{1}{n}\sum_{i=1}^n Y_i$, under the proposed policies of each approach. Each row corresponds to the average value for a single iteration of the specified simulation setup.
    \item \texttt{all\_sims\_cont\_outcomes.csv}: This file contains the average continuous outcome value under the proposed policies of each approach. The continuous outcome is simply $O_i$ rather than $Y_i$ in our data generation process outlined in Appendix~\ref{sec: dgp}. We can report these values since we know the true underlying data generation process. Each row corresponds to the average value for a single iteration of the specified simulation setup.
    \item
    \texttt{all\_sims\_nan.csv}: This file contains the number of iterations that each approach failed to produce policy estimates for the 32 simulation setups. See details in Appendix~\ref{sec: appendix-add-synth-details} for explanation on why methods may have failed.
    \item 
    \texttt{sims\_binary\_outcomes\_mean.csv}: This file contains the average binary outcome value across all iterations of each simulation setup for each method. Note that not all methods ran for 20 iterations for each setup. See \texttt{all\_sims\_nan.csv}.
    \item
    \texttt{sims\_binary\_outcomes\_std.csv}: This file contains the standard deviation of the average binary outcome value across all iterations of each simulation setup for each method. Note that not all methods ran for 20 iterations for each setup. See \texttt{all\_sims\_nan.csv}.
    
    \item\texttt{sims\_binary\_outcomes\_median.csv}: This file contains the median of the average binary outcome value across all iterations of each simulation setup for each method. Note that not all methods ran for 20 iterations for each setup. See \texttt{all\_sims\_nan.csv}.
\end{itemize}
We also include \texttt{sims\_cont\_outcomes\_mean.csv}, \texttt{sims\_cont\_outcomes\_std.csv}, and \texttt{sims\_cont\_outcomes\_median.csv} which contain the same content but for the continuous outcome.
\section{\textsc{Data Summary}}\label{sec: data_desc}
\begin{longtable}{ll}
\caption{Full cohort characteristics and data description.}\\
\hline
\quad\quad\textbf{Variable} & \quad\textbf{Value} \\ \hline
Age, year, median (IQR) & 61 (48 -- 73) \\ \hline
Male gender, n (\%) & 475 (47.7\%) \\ \hline
Race &  \\ \hline
\quad Asian, n (\%) & 33 (3.3\%) \\ \hline
\quad Black / African American, n (\%) & 72 (7.2\%) \\ \hline
\quad White / Caucasian, n (\%) & 751 (75.5\%) \\ \hline
\quad Other, n (\%) & 50 (5.0\%) \\ \hline
\quad Unavailable / Declined, n (\%) & 84 (8.4\%) \\ \hline
Married, n (\%) & 500 (50.3\%) \\ \hline
Premorbid mRS before admission, median (IQR) & 0 (0 -- 3) \\ \hline
APACHE II in first 24h, median (IQR) & 19 (11 -- 25) \\ \hline
Initial GCS, median (IQR) & 11 (6 -- 15) \\ \hline
Initial GCS is with intubation, n (\%) & 415 (41.7\%) \\ \hline
Worst GCS in first 24h, median (IQR) & 8 (3 -- 14) \\ \hline
Worst GCS in first 24h is with intubation, n (\%) & 511 (51.4\%) \\ \hline
Admitted due to surgery, n (\%) & 168 (16.9\%) \\ \hline
Cardiac arrest at admission, n (\%) & 79 (7.9\%) \\ \hline
Seizure at presentation, n (\%) & 228 (22.9\%) \\ \hline
Acute SDH at admission, n (\%) & 146 (14.7\%) \\ \hline
Take anti-epileptic drugs outside hospital, n (\%) & 123 (12.4\%) \\ \hline
Highest heart rate in first 24h, /min, median (IQR) & 92 (80 -- 107) \\ \hline
Lowest heart rate in first 24h, /min, median (IQR) & 71 (60 -- 84) \\ \hline
Highest systolic BP in first 24h, mmHg, median (IQR) & 153 (136 -- 176) \\ \hline
Lowest systolic BP in first 24h, mmHg, median (IQR) & 116 (100 -- 134) \\ \hline
Highest diastolic BP in first 24h, mmHg, median (IQR) & 84 (72 -- 95) \\ \hline
Lowest diastolic BP in first 24h, mmHg, median (IQR) & 61 (54 -- 72) \\ \hline
Mechanical ventilation on the first day of EEG, n (\%) & 572 (57.5\%) \\ \hline
Systolic BP on the first day of EEG, mmHg, median (IQR) & 148 (130 -- 170) \\ \hline
GCS on the first day of EEG, median (IQR) & 8 (5 -- 13) \\ \hline
History &  \\ \hline
\quad Stroke, n (\%) & 192 (19.3\%) \\ \hline
\quad Hypertension, n (\%) & 525 (52.8\%) \\ \hline
\quad Seizure or epilepsy, n (\%) & 182 (18.3\%) \\ \hline
\quad Brain surgery, n (\%) & 109 (11.0\%) \\ \hline
\quad Chronic kidney disorder, n (\%) & 112 (11.3\%) \\ \hline
\quad Coronary artery disease and myocardial infarction, n (\%) & 160 (16.1\%) \\ \hline
\quad Congestive heart failure, n (\%) & 90 (9.0\%) \\ \hline
\quad Diabetes mellitus, n (\%) & 201 (20.2\%) \\ \hline
\quad Hypersensitivity lung disease, n (\%) & 296 (29.7\%) \\ \hline
\quad Peptic ulcer disease, n (\%) & 50 (5.0\%) \\ \hline
\quad Liver failure, n (\%) & 46 (4.6\%) \\ \hline
\quad Smoking, n (\%) & 461 (46.3\%) \\ \hline
\quad Alcohol abuse, n (\%) & 231 (23.2\%) \\ \hline
\quad Substance abuse, n (\%) & 119 (12.0\%) \\ \hline
\quad Cancer (except central nervous system), n (\%) & 180 (18.1\%) \\ \hline
\quad Central nervous system cancer, n (\%) & 85 (8.5\%) \\ \hline
\quad Peripheral vascular disease, n (\%) & 41 (4.1\%) \\ \hline
\quad Dementia, n (\%) & 45 (4.5\%) \\ \hline
\quad Chronic obstructive pulmonary disease or asthma, n (\%) & 139 (14.0\%) \\ \hline
\quad Leukemia or lymphoma, n (\%) & 22 (2.2\%) \\ \hline
\quad AIDS, n (\%) & 12 (1.2\%) \\ \hline
\quad Connective tissue disease, n (\%) & 47 (4.7\%) \\ \hline
Primary diagnosis &  \\ \hline
\quad Septic shock, n (\%) & 131 (13.2\%) \\ \hline
\quad Ischemic stroke, n (\%) & 85 (8.5\%) \\ \hline
\quad Hemorrhagic stroke, n (\%) & 163 (16.4\%) \\ \hline
\quad Subarachnoid hemorrhage (SAH), n (\%) & 188 (18.9\%) \\ \hline
\quad Subdural hematoma (SDH), n (\%) & 94 (9.4\%) \\ \hline
\quad SDH or other traumatic brain injury including SAH, n (\%) & 52 (5.2\%) \\ \hline
\quad Traumatic brain injury including SAH, n (\%) & 21 (2.1\%) \\ \hline
\quad Seizure/status epilepticus, n (\%) & 258 (25.9\%) \\ \hline
\quad Brain tumor, n (\%) & 113 (11.4\%) \\ \hline
\quad CNS infection, n (\%) & 64 (6.4\%) \\ \hline
\quad Ischemic encephalopathy or Anoxic brain injury, n (\%) & 72 (7.2\%) \\ \hline
\quad Toxic metabolic encephalopathy, n (\%) & 104 (10.5\%) \\ \hline
\quad Primary psychiatric disorder, n (\%) & 35 (3.5\%) \\ \hline
\quad Structural-degenerative diseases, n (\%) & 35 (3.5\%) \\ \hline
\quad Spell, n (\%) & 5 (0.5\%) \\ \hline
\quad Respiratory disorders, n (\%) & 304 (30.6\%) \\ \hline
\quad Cardiovascular disorders, n (\%) & 153 (15.4\%) \\ \hline
\quad Kidney failure, n (\%) & 65 (6.5\%) \\ \hline
\quad Liver disorder, n (\%) & 30 (3.0\%) \\ \hline
\quad Gastrointestinal disorder, n (\%) & 18 (1.8\%) \\ \hline
\quad Genitourinary disorder, n (\%) & 34 (3.4\%) \\ \hline
\quad Endocrine emergency, n (\%) & 28 (2.8\%) \\ \hline
\quad Non-head trauma, n (\%) & 13 (1.3\%) \\ \hline
\quad Malignancy, n (\%) & 65 (6.5\%) \\ \hline
\quad Primary hematological disorder, n (\%) & 24 (2.4\%) \\ \hline
\label{tab:Ctab}
\end{longtable}
\section{\textsc{Anti-Seizure Medications and Policy Templates}}\label{sec: policy}

\subsection{Anti-Seizure Medications}
\label{subsec:asm_info}
\begin{figure}
    \centering
    \includegraphics[width=\textwidth]{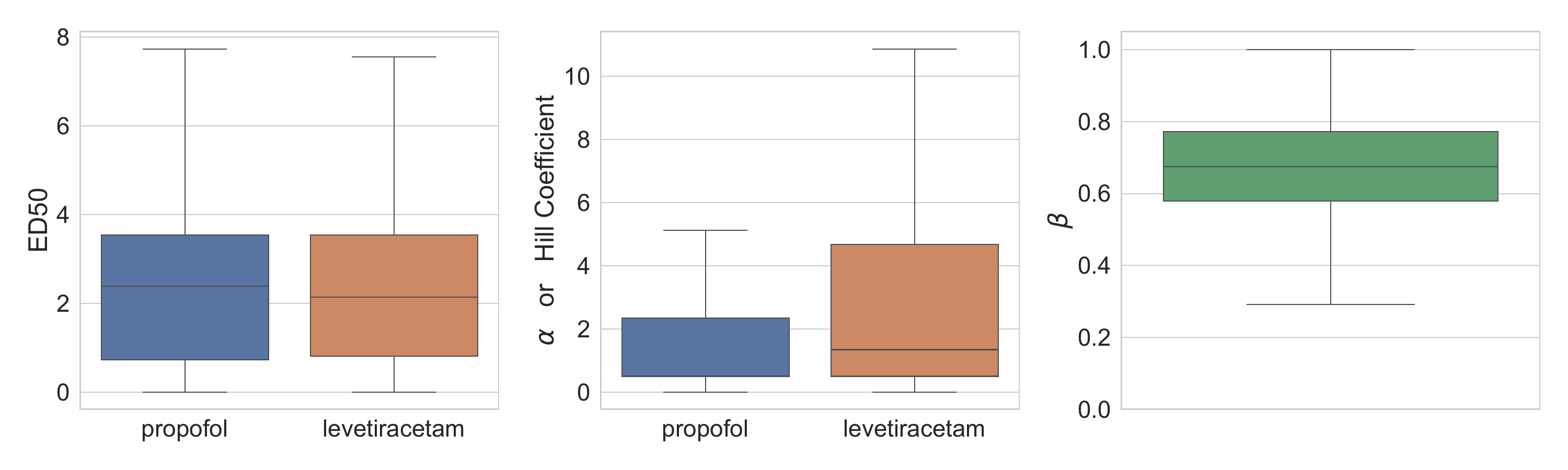}
    \caption{Boxplots showing the distribution of the estimated pharmacodynamics parameters.}
    \label{fig:pd_est}
\end{figure}

Two drugs were studied: propofol and levetiracetam, Propofol is a sedative antiseizure medication and is given as a continuous infusion, while levetiracetam is a non-sedative antiseizure medication given as a bolus. The doses are normalized by body weight (kg). We use the half-lives from the literature for estimating the drug concentrations $\D$ and estimate the PD parameters using the $E$ and $\D$ for each patient in our cohort (see Table~\ref{tab:drugcx} and Figure~\ref{fig:pd_est}). 
\begin{table}[h!]
\centering
\caption{PK and the estimated average PD parameters for the anti-seizure medications. }
{
\begin{tabular}{cccc}
\hline
\textbf{Drug}  & \textbf{Half-Life}  & \textbf{avg. $\widehat{ED50}$}&$\mathbf{avg. \widehat{\alpha}}$\\
\hline
Propofol & 20 minutes  & 2.41 mg/kg/hr&2.96\\
Levetiracetam & 8 hours  & 2.26 mg/kg&3.33\\
\hline
\end{tabular}}
\label{tab:drugcx}
\end{table}

\subsection{Policy Templates}\label{sec:appdx-temp}
The regime determining the dose for patient $i$, for propofol at time $t$ is given by:
{\small \begin{eqnarray}\label{eq: def_pol_prop}
&&\pi^{prop}_{i}\left(\{E_{i,t'}\}_{t'=1}^{t-1},\{\Z_{i,t'}\}_{t'=1}^{t-1} ; \ba^{p}_i \right) \nonumber \\
    &&= a^{p}_{1,i} \mathbf{1}[ E_{i,t-1hr} > 25\% ] + a^{p}_{2,i} \mathbf{1}[ E_{i,t-1hr} > 50\% ] \nonumber \\
    &&+ a^{p}_{3,i} \mathbf{1}[ E_{i,t-1hr} > 75\% ] \nonumber \\ &&+ a^{p}_{4,i} \mathbf{1}[ E_{i,t-6hr} > 25\% ]
    + a^{p}_{5,i} \mathbf{1}[ E_{i,t-6hr} > 50\% ] \nonumber\\ && + a^{p}_{6,i} \mathbf{1}[ E_{i,t-1hr} > 25\% ] \mathbf{1}[ E_{i,t-6hr} > 25\% ] \nonumber \\ && + a^{p}_{7,i}\mathbf{1}[ E_{i,t-6hr} > 25\% ] \mathbf{1}[ E_{i,t-12hr} > 25\% ],
\end{eqnarray}}
where $\ba^{p}$ is a vector of the parameters for propofol's regime, $E_{i,t-t'}$ is the average EA burden between time $t-t'$ and $t$, and $Z_{i,j',t-t'}$ is the total dose of drug $j'$ administered between time $t-t'$ and $t$. 

The regime determining the dose for patient $i$, for levetiracetam at time $t$ is given by:
{\small \begin{eqnarray}\label{eq: def_pol_lev}
&&\pi^{lev}_{i}\left(\{E_{i,t'}\}_{t'=1}^{t-1},\{\Z_{i,t'}\}_{t'=1}^{t-1} ; \ba^{l}_i \right) \nonumber \\
    &&= \mathbf{1}\left[Z_{\text{lev},i,t-12hr} = 0\right] \times \nonumber \\ && \bigg( a^{l}_{0,i} + a^{l}_{1,i} \mathbf{1}[ E_{i,t-1hr} > 25\% ] \nonumber \\ &&+ a^{l}_{2,i} \mathbf{1}[ E_{i,t-1hr} > 50\% ] + a^{l}_{3,i} \mathbf{1}[ E_{i,t-1hr} > 75\% ] \nonumber \\ &&+ a^{l}_{4,i} \mathbf{1}[ E_{i,t-6hr} > 25\% ]
    + a^{l}_{5,i} \mathbf{1}[ E_{i,t-6hr} > 50\% ] \nonumber\\ && + a^{l}_{6,i} \mathbf{1}[ E_{i,t-1hr} > 25\% ] \mathbf{1}[ E_{i,t-6hr} > 25\% ] \nonumber \\ && + a^{l}_{7,i}\mathbf{1}[ E_{i,t-6hr} > 25\% ] \mathbf{1}[ E_{i,t-12hr} > 25\% ]\bigg),
\end{eqnarray}}
where $\ba^{l}$ is a vector of the parameters for levetiracetam's regime, 

Thus, the regime for patient $i$, denoted by $$\pi_i = \begin{Bmatrix} 
\pi^{prop}_{i}\left(\{E_{i,t'}\}_{t'=1}^{t-1},\{\Z_{i,t'}\}_{t'=1}^{t-1} ; \ba^{p}_i \right) \\
\pi^{lev}_{i}\left(\{E_{i,t'}\}_{t'=1}^{t-1},\{\Z_{i,t'}\}_{t'=1}^{t-1} ; \ba^{l}_i \right)\}
\end{Bmatrix}$$
We estimate $\ba^{p}$ and $\ba^{l}$ by minimizing the mean squared error loss between the predicted drug doses and the observed drug doses, $Z_{\text{prop}, i, t}$ and $Z_{\text{lev}, i, t}$ at each time $t$.


\section{\textsc{Consistency Proposition and Proof}}\label{sec: appendix_theorem}
Before proceeding to the proof of Proposition~\ref{prop:consistent}, we note that we consider optimality with respect to linear score functions inside the convex hull of locally observed policies. Our methodology is versatile but operationalized with a linear score function that aligns with the policy template of a prominent tertiary hospital we target (details in Appendix~\ref{sec:appdx-temp}). In this context, our approach identifies an optimal treatment regime aimed at minimizing the probability of adverse outcomes, such as death. Emphasizing patient safety, our search is confined within the convex hull of observed policies for “similar” patients. We now present the proof for Proposition~\ref{prop:consistent}.

\begin{propositionrepeat}[Consistency of Treatment Regime Estimator]
    Consider a nest sequence of datasets $\{\mathcal{D}_n\}$ such that $|\mathcal{D}_n| = n$. Then, given conditional ignorability, local positivity, and the smooth outcomes assumptions, 
    \begin{equation*}
        \lim_{n\to\infty} \mathbb{E}[Y_i(\widehat{\pi}^{*,(n)}_i) \mid \V_i] \to \mathbb{E}[Y_i({\pi}^*_i) \mid \V_i] ,
    \end{equation*}
    where $\widehat{\pi}^{*,(n)}_i$ is the estimate of the optimal treatment regime for unit $i$ estimated using the caliper nearest neighbors interpolation on dataset $\mathcal{D}_n$ with caliper $r_n$. 
\end{propositionrepeat}
\paragraph{Proof.} 
Let $\mu_i(\bv,\pi) :=  \E[ Y_i(\pi) \mid \V_i = \bv]$ be the expected potential outcome for unit $i$ for which we are interested in estimating the optimal policy, and 
\begin{eqnarray*}
    A^{(n)}_i := \mu_i(\V_i,\widehat{\pi}^{*,(n)}_i) - \mu_i(\V_i,\pi^*_i).
\end{eqnarray*}
By conditional ignorability, $\mu_i(\bv,\pi) = \E[ Y_i \mid \V_i = \bv, \pi_i = \pi]$. Also, let $\widehat{\mu}^{(n)}_i(\bv,\pi)$ denote the $r_n$-caliper nearest neighbor estimate of $\mu_i(\bv,\pi)$ on dataset $\mathcal{D}_n$ and $MG^{(n)}_i$ denote the set of all units in $\mathcal{D}_n$ that are at max $r_n$ distance away from $\V_i$. Then, by definition, $\pi^*_i$ is the policy that, given $\V_i$, minimizes $\mu_i(\cdot,\cdot)$, and $\widehat{\pi}^{*,(n)}_i$ is the policy that, given $\V_i$, minimizes $\widehat{\mu}^{(n)}_i(\cdot,\cdot)$. Thus,
\begin{eqnarray*}
    A^{(n)}_i &\leq& \left( \mu_i(\V_i,\widehat{\pi}^{*,(n)}_i) - \widehat{\mu}^{(n)}_i(\V_i,\widehat{\pi}^{*,(n)}_i) \right) - \left( \mu_i(\V_i,\pi^*_i) - \widehat{\mu}^{(n)}_i(\V_i,\pi^*_i) \right) \\
    &\leq& \left| \left( \mu_i(\V_i,\widehat{\pi}^{*,(n)}_i) - \widehat{\mu}^{(n)}_i(\V_i,\widehat{\pi}^{*,(n)}_i) \right) - \left( \mu_i(\V_i,\pi^*_i) - \widehat{\mu}^{(n)}_i(\V_i,\pi^*_i) \right) \right| \\
     &\leq& \left| \left(\mu_i(\V_i,\pi^*_i) - \widehat{\mu}^{(n)}_i(\V_i,\pi^*_i) \right) \right| + \left|\left( \mu_i(\V_i,\widehat{\pi}^{*,(n)}_i) - \widehat{\mu}^{(n)}_i(\V_i,\widehat{\pi}^{*,(n)}_i) \right) \right|.
\end{eqnarray*}

As $n \to \infty$ we shrink $r_n \to 0$ such that $|MG^{(n)}_i| \to \infty$. Then, by the consistency of the caliper nearest-neighbors estimator under smoothness of outcomes, $\widehat{\mu}^{(n)}_i(\V_i,\pi) \to \mu_i(\V_i,\pi)$ (see Remark~\ref{remark: consistent}). This implies that, as $n \to \infty$, $A^{(n)} = \mu_i(\V_i,\widehat{\pi}^{*,(n)}_i) - \mu_i(\V_i,\pi^*_i) \to A^{(\infty)} \leq 0$. Further, by definition of , $\mu_i(\V_i,\pi^*_i) \leq \mu_i(\V_i,\widehat{\pi}^{*,(n)}_i)$. Thus, we get, $\mu_i(\V_i,\widehat{\pi}^{*,(n)}_i) \to \mu_i(\V_i,\pi^*_i)$, as $n \to \infty$. \textbf{QED.}

\begin{remark}\label{remark: consistent}
    The consistency of the caliper nearest-neighbors estimator is a standard and well-explored result in the literature \citep{parikh2022malts, devroye1994strong, kudraszow2013uniform, li1984consistency, jiang2019non, ferraty2010rate, kara2017data, einmahl2005uniform}. Our context is similar to the one discussed in Theorem 1 of \citet{parikh2022malts} and Theorem 2 of \citet{kudraszow2013uniform}. 
\end{remark}

\begin{remark} 
    The results in Theorem 2.2 from \cite{zhou2017causal}, shows similar consistency result of the optimal treatment regime estimator.
\end{remark}


\end{appendices}
\end{document}